\documentclass[lettersize,journal]{IEEEtran}
\usepackage{amsmath,amsfonts}
\usepackage{algorithmic}
\usepackage{algorithm}
\usepackage{array}
\usepackage[caption=false,font=normalsize,labelfont=sf,textfont=sf]{subfig}
\usepackage{textcomp}
\usepackage{stfloats}
\usepackage{url}
\usepackage{verbatim}
\usepackage{graphicx}
\usepackage{cite}
\usepackage{amssymb}
\usepackage{multirow}
\usepackage{booktabs}
\usepackage{array}
\usepackage{enumitem}
\usepackage{color}
\usepackage{tabularx}
\usepackage{setspace}

\newcommand{\etal}{\textit{et al}., }
\newcommand{\ie}{\textit{i}.\textit{e}., }
\newcommand{\eg}{\textit{e}.\textit{g}., }
\newcommand{\aka}{\textit{a}.\textit{k}.\textit{a}., }

\begin{document}

\title{Deep Learning-based Occluded Person Re-identification: A Survey}

\author{
Yunjie~Peng,~Saihui~Hou,~Chunshui~Cao,~Xu~Liu,\\~Yongzhen~Huang,~Zhiqiang~He\IEEEauthorrefmark{1}% <-this % stops a space
\thanks{\IEEEauthorrefmark{1}Corresponding author.}% <-this % stops a space
\thanks{Yunjie Peng is with the School of Computer Science and Technology, Beihang University, Beijing 100191, China (email: yunjiepeng@buaa.edu.cn).}% <-this % stops a space
\thanks{Saihui Hou and Yongzhen Huang are with the School of Artificial Intelligence, Beijing Normal University, Beijing 100875, China and also with Watrix Technology Limited Co. Ltd, Beijing, China (email: housaihui@bnu.edu.cn; huangyongzhen@bnu.edu.cn).}% <-this % stops a space
\thanks{Chunshui Cao and Xu Liu are with the Watrix Technology Limited Co. Ltd, Beijing, China (email: chunshui.cao@watrix.ai; xu.liu@watrix.ai).}% <-this % stops a space
\thanks{Zhiqiang He is with the School of Computer Science and Technology, Beihang University, Beijing 100191, China and the Lenovo Corporation, Beijing, China (email: zqhe1963@gmail.com).}}

\markboth{Journal of \LaTeX\ Class Files,~Vol.~14, No.~8, August~2021}%
{Shell \MakeLowercase{\textit{et al.}}: A Sample Article Using IEEEtran.cls for IEEE Journals}


\maketitle

\begin{abstract}
Occluded person re-identification (Re-ID) aims at addressing the occlusion problem when retrieving the person of interest across multiple cameras.
With the promotion of deep learning technology and the increasing demand for intelligent video surveillance, the frequent occlusion in real-world applications has made occluded person Re-ID draw considerable interest from researchers.
A large number of occluded person Re-ID methods have been proposed while there are few surveys that focus on occlusion.
To fill this gap and help boost future research, this paper provides a systematic survey of occluded person Re-ID.
Through an in-depth analysis of the occlusion in person Re-ID, most existing methods are found to only consider part of the problems brought by occlusion.
Therefore, we review occlusion-related person Re-ID methods from the perspective of issues and solutions.
We summarize four issues caused by occlusion in person Re-ID, \ie position misalignment, scale misalignment, noisy information, and missing information.
The occlusion-related methods addressing different issues are then categorized and introduced accordingly.
After that, we summarize and compare the performance of recent occluded person Re-ID methods on four popular datasets: Partial-ReID, Partial-iLIDS, Occluded-ReID, and Occluded-DukeMTMC.
Finally, we provide insights on promising future research directions.
% \textcolor{blue}{
% With the promotion of deep learning technology and the increasing demand for intelligent video surveillance, it has drawn considerable interest from researchers for its real-world applications.
% %
% Occluded person Re-ID has been widely studied and a large number of methods have been proposed.
% %
% In the literature, there are many surveys on person Re-ID but few of them focus on occlusion.
% %
% Considering the complex and diverse solutions in the field, our systematic survey is essential to help boost future research.
% %
% Specifically, we review methods from the perspective of issues and solutions.
% %
% Through an in-depth analysis of the occlusion in person Re-ID, existing methods are found to only consider part of the problems caused by occlusion.
% %
% Therefore, we first summarize the four inherent issues in occluded person Re-ID, \ie position misalignment, scale misalignment, noisy information, and missing information.
% %
% The occlusion-related methods addressing different issues are then categorized and introduced accordingly.
% %
% We summarize and compare the performance of recent occluded person Re-ID methods on four popular Re-ID datasets: Partial-ReID, Partial-iLIDS, Occluded-ReID, and Occluded-Duke.
% %
% Finally, we provide insights on promising future research directions.}
% %
\end{abstract}

\begin{IEEEkeywords}
Occluded Person Re-identification, Partial Person Re-identification, Literature Survey, Deep Learning.
\end{IEEEkeywords}

\section{Introduction}
%What is Person Re-ID and why is it important.
\IEEEPARstart{P}{erson} re-identification (Re-ID) retrieves persons of the same identity across different cameras~\cite{ye2021tpamisurvey}.
With the expanding deployment of surveillance cameras and the increasing demand for public safety, person Re-ID which plays the fundamental role in intelligent surveillance has become a research hotspot in the computer vision community.
%Challenges in person Re-ID.
In practice, the internal variations of a person (\eg pose variations~\cite{zhao2017spindle,sarfraz2018pose,liu2018pose} and clothes changes~\cite{qian2020long,hong2021fine}), as well as the complex environments (\eg illumination changes~\cite{huang2019illumination,zhang2022illumination}, viewpoint variations~\cite{karanam2015viewpoint,sun2019dissecting,wu2020multiview}, and occlusion~\cite{miao2021visible,chen2021occlude,hou2021completion}), bring significant challenges to person Re-ID.
%The reason why the occluded person Re-ID is important and is separately proposed.
Among them, the occlusion which occurs frequently in real-world applications and affects the accuracy greatly has received considerable interest from researchers.

Occluded person re-identification~\cite{zheng2015partial,zhuo2018occluded,hou2021completion} is proposed to address the occlusion problem for real-world person re-identification.
Different from general person Re-ID approaches which assume the retrieval process with whole human body available, occluded person Re-ID highlights the scenario of pedestrian images occluded by various obstacles (\eg cars, trees, and crowds) and focuses on retrieving persons of the same identity when given an occluded query.
\begin{figure}  
\centering  
\includegraphics[width=0.47\textwidth]{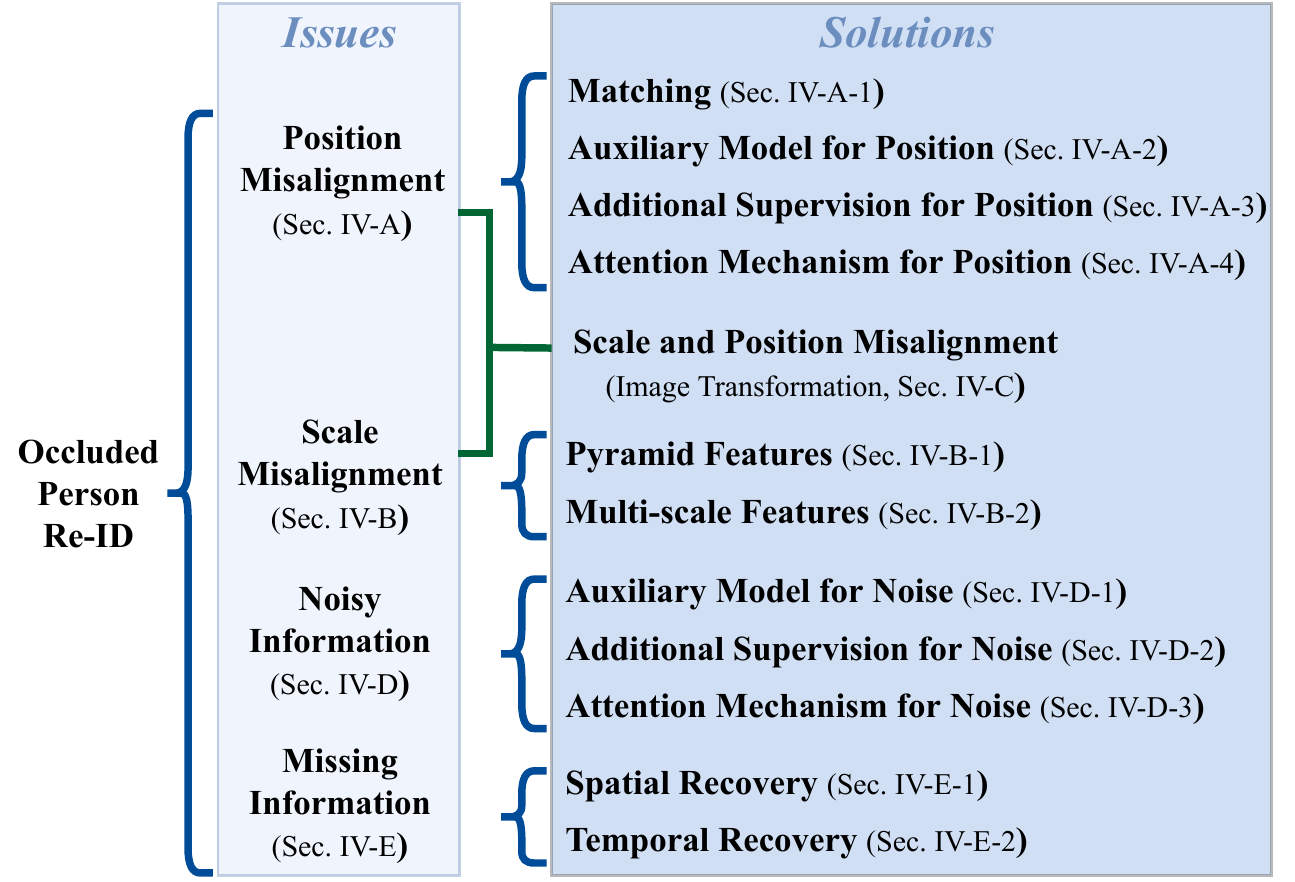}
\caption{The taxonomy of occluded person Re-ID methods from the perspective of issues and solutions.
Using the above taxonomy, it is easy to know the inherent challenges for occluded person Re-ID and have a general understanding of overall technical routes.
}\label{fig_taxonomy}
\end{figure}

\begin{figure}  
\centering  
\includegraphics[width=0.44\textwidth]{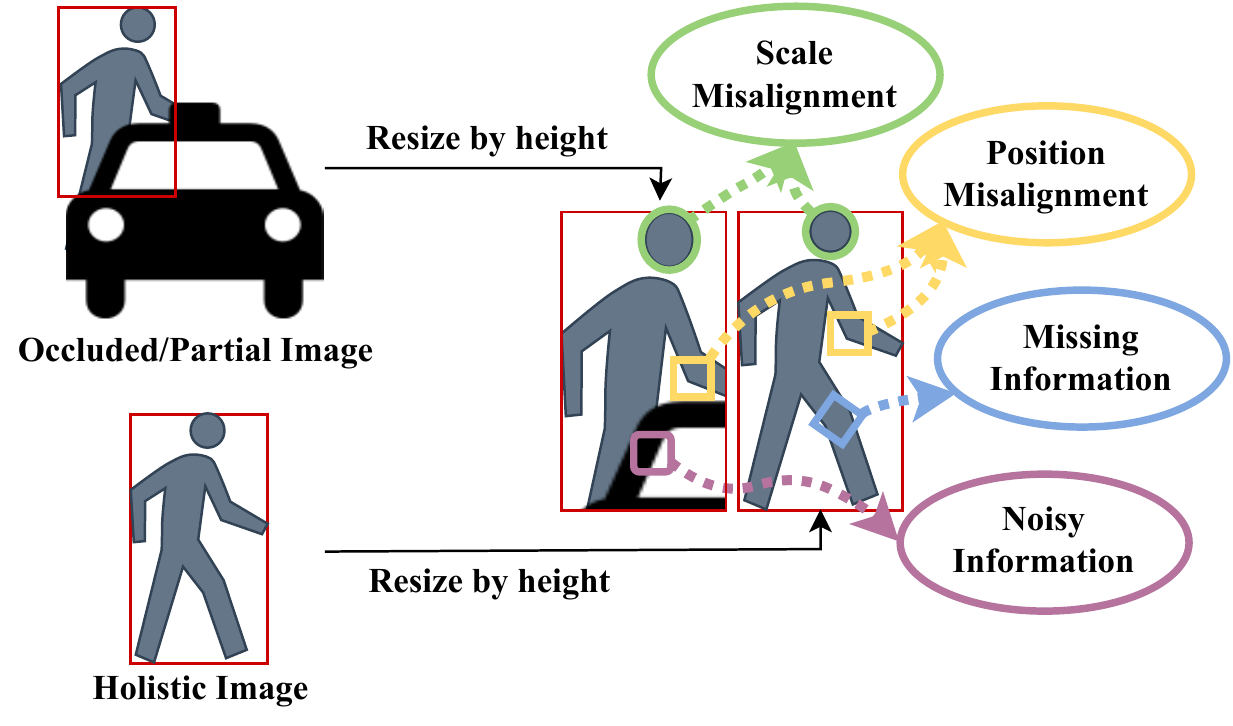}
\caption{The position misalignment, scale misalignment, noisy information, and missing information issues caused by occlusion for person Re-ID.}
\label{fig_four_issues}
\end{figure}

With the advancement of deep learning, a large number of occluded person Re-ID methods have been proposed while there are few surveys that focus on occlusion.
To fill this gap, this paper summarizes occlusion-related person Re-ID works and provides a systematic survey of occluded person Re-ID.
Through an in-depth analysis of the occlusion in person re-identification, most existing methods are found to only consider part of the problems caused by occlusion.
Therefore, we review occluded person Re-ID from the perspective of issues and solutions to facilitate the understanding of the latest approaches and inspire new ideas in the field.
The issues caused by occlusion for person Re-ID are carefully summarized from the whole process of person re-identification.
Technically speaking, a practical person Re-ID system in video surveillance mainly consists of three stages~\cite{zheng2016survey}: pedestrian detection, trajectory tracking, and person retrieval.
Although it is generally believed that the first two stages are independent computer vision tasks and most person Re-ID works focus on person retrieval, the occlusion will affect the whole process and bring great challenges to the final re-identification.
In summary, four significant issues are considered for occluded person Re-ID in this paper: the position misalignment, the scale misalignment, the noisy information, and the missing information.
Each issue is illustrated in Fig.~\ref{fig_four_issues} and we briefly introduce each issue as follows.

1). \emph{The position misalignment.}
Generally, the detected human boxes are resized by height to obtain the same size of input data for person retrieval.
In the case of occlusion, the detected box of the person contains only part of the human body while it undergoes the same alignment processing as that of a non-occluded person.
The contents at the same position of the processed partial image and the processed holistic image are likely to be mismatched, resulting in the position misalignment issue.
2). \emph{The scale misalignment.}
Similar to the position misalignment, the scale misalignment also arises from the upstream data processing procedure.
The occlusion may affect the height of the detected box and thus influence the resizing ratio in the data processing, resulting in the scale misalignment between a partial and a holistic image.
3). \emph{The noisy information.}
In the detected boxes of occluded pedestrians, the occlusion is inevitably included in whole or in part and brings the noisy information for person Re-ID.
4). \emph{The missing information.}
In the detected boxes of occluded pedestrians, the identity information of occluded regions is missing, resulting in the missing information issue.

This paper analyzes occlusion-related person Re-ID methods regarding the above-mentioned four issues and provides a multi-dimensional taxonomy to categorize solutions for each issue (see Fig.~\ref{fig_taxonomy}).
Specifically, we mainly review published publications of deep learning-based occluded person Re-ID from top conferences and journals before June, 2022, and meanwhile we also introduce some methods from other conferences and journals as supplements.
We discuss the issues brought by occlusion for person Re-ID and provide an in-depth analysis of how the issues are addressed in recent works with evaluation results summarized accordingly.
Particularly, some person Re-ID methods are closely related to occlusion and we also summarize these methods to obtain a more comprehensive survey for occluded person Re-ID.
The main contributions of this survey lie in three aspects:
\begin{itemize}
\item{To fill the gap of the occluded person Re-ID survey, we review recent person re-identification methods for occlusion and provide a systematic survey from the perspective of issues and solutions.}
\item{We summarize and compare the performance of mainstream occluded person Re-ID approaches for researchers and industries to use based on their practical needs.}
\item{We summarize and analyze the advantages and disadvantages of different types of solutions for occluded person Re-ID and provide insights on promising research directions in this field.}
\end{itemize}

The rest of this paper is organized as follows.
Section 2 presents a summary of previous surveys and elaborates on efforts made by this survey compared with others.
Section 3 summarizes the common datasets and evaluation metrics of occluded person Re-ID.
Section 4 provides an in-depth analysis of deep learning-based occluded person Re-ID methods from the perspective of issues and solutions.
Section 5 compares the performance of various solutions and provides insights on promising research directions.
Section 6 gives our conclusions.

\section{Previous Surveys}
In the previous literature, there are some surveys that have also reviewed the field of person Re-ID.
To obtain a more comprehensive comparison, we summarize the surveys of person Re-ID since 2012.
The taxonomies of these surveys are listed in Table~\ref{table_previous_survey}.
On the whole, previous surveys of person Re-ID can be roughly divided into traditional surveys~\cite{mazzon2012survey,bedagkar2014survey} and deep learning-based surveys~\cite{zheng2016survey,lavi2018survey,wu2019survey,leng2019survey,lavi2020survey,karanam2019survey,islam2020survey,lin2021survey,lin2021survey2,ye2021tpamisurvey,ming2022survey}.

Traditional surveys mainly review person Re-ID methods that manually design the feature extraction procedure and learn a better similarity measurement.
Mazzon \etal\cite{mazzon2012survey} summarize four main phases for person Re-ID: multi-person detection, feature extraction, cross-camera calibration, and person association.
Assuming that the first phase (\ie multi-person detection) has been solved, methods of extracting color/texture/shape appearance features, grouping temporal information, conducting color/spatio-temporal cross-camera calibration, and distance/learning/optimization-based person association are reviewed accordingly.
Apurva \etal\cite{bedagkar2014survey} regard the tracking across multiple cameras as the open set matching problem and the identity retrieval as the close set matching problem. 
According to whether additional camera geometry/calibration information is available, methods are divided into contextual Re-ID and non-contextual Re-ID for open-set and close-set matching respectively.

Deep learning-based surveys mainly summarize the person Re-ID methods using deep learning techniques from different perspectives.
A few of these surveys~\cite{zheng2016survey,karanam2019survey} also review traditional methods for the sake of completeness.
In general, previous deep learning-based surveys have involved loss design~\cite{lavi2018survey,islam2020survey}, technical means~\cite{islam2020survey,ming2022survey}, data augmentation~\cite{lavi2018survey,wu2019survey}, image and video~\cite{zheng2016survey,ye2021tpamisurvey}, classification and verification~\cite{lavi2018survey,wu2019survey,lavi2020survey}, open-world and close-world~\cite{leng2019survey,ye2021tpamisurvey}, multi-modality~\cite{leng2019survey,lin2021survey2}, ranking optimization~\cite{karanam2019survey,ye2021tpamisurvey}, noisy annotation~\cite{ye2021tpamisurvey}, unsupervised learning~\cite{lin2021survey}, and metric learning~\cite{wu2019survey,karanam2019survey,islam2020survey,ming2022survey,ye2021tpamisurvey}.
Despite such a number of surveys on person Re-ID, the occlusion problem has not drawn enough attention and only a few surveys pay attention to occluded person Re-ID.
As far as we know, Ye \etal~\cite{ye2021tpamisurvey} have made a rough summary of occluded person Re-ID as a part of Noise-Robust Re-ID.
Ming \etal~\cite{ming2022survey} have included several occluded person Re-ID methods in local feature learning and sequence feature learning.
Considering the practical importance of occlusion for person Re-ID, a systematic investigation for occluded person Re-ID is essential.
Therefore, we provide an in-depth survey of issues and solutions involved in occlusion-related person Re-ID works to help boost future research.
\begin{table}[!t]
\centering
\caption{The Overview of Person Re-ID Surveys in Recent Years.\label{table_previous_survey}}
\setlength \tabcolsep{1.4pt}
\begin{spacing}{1.15}
\resizebox{0.98\linewidth}{!}{%
\begin{tabular}{cll}
\toprule
\multicolumn{1}{c}{Surveys} & 
\multicolumn{1}{c}{Reference} & 
\multicolumn{1}{c}{Taxonomy} \\ \cline{1-3}

\multicolumn{1}{c}{\multirow{4}{*}{Traditional}} &
2012 PRL~\cite{mazzon2012survey} & 
\multicolumn{1}{l}{
\begin{tabular}{l}
Feature Extraction / Cross-camera Calibration /\\ Person Association
\end{tabular}} \\ \cline{2-3}

\multicolumn{1}{c}{} &
2014 IVC~\cite{bedagkar2014survey} & 
\multicolumn{1}{l}{
\begin{tabular}{l}
Contextual Methods (camera geometry info;\\ camera calibration.) / Non-contextual Methods\\ (passive methods; active methods)
\end{tabular}} \\ \cline{1-3}

\multicolumn{1}{c}{\multirow{29}{*}{\shortstack{Deep\\learning-based}}} &
2016 arXiv~\cite{zheng2016survey} & 
\multicolumn{1}{l}{
\begin{tabular}{l}
Image-based Methods / Video-based Methods
\end{tabular}} \\ \cline{2-3}

\multicolumn{1}{c}{} &
2019 TCSVT~\cite{leng2019survey} & 
\multicolumn{1}{l}{
\begin{tabular}{l}
Person Verification / Application-driven Methods\\ (raw data; practial procedure; efficiency)
\end{tabular}} \\ \cline{2-3}

\multicolumn{1}{c}{} &
2019 TPAMI~\cite{karanam2019survey} & 
\multicolumn{1}{l}{
\begin{tabular}{l}
Feature Extraction / Metric Learning / Multi-shot\\ Ranking
\end{tabular}} \\ \cline{2-3}

\multicolumn{1}{c}{} &
2020 arXiv~\cite{lavi2020survey} & 
\multicolumn{1}{l}{
\begin{tabular}{l}
Identification Task / Verification Task
\end{tabular}} \\ \cline{2-3}

\multicolumn{1}{c}{} &
2020 IVC~\cite{islam2020survey} & 
\multicolumn{1}{l}{
\begin{tabular}{l}
Feature Learning / Model Architecture Design /\\ Metric Learning / Loss Function 
\end{tabular}} \\ \cline{2-3}

\multicolumn{1}{c}{} &
2021 arXiv~\cite{lin2021survey} & 
\multicolumn{1}{l}{
\begin{tabular}{l}
Pseudo-label Estimation / Deep Feature\\ Representation Learning / Camera-aware\\ Invariance Learning / Unsupervised Domain\\ Adaptation
\end{tabular}} \\ \cline{2-3}

\multicolumn{1}{c}{} &
2021 IJCAI~\cite{wang2021survey} & 
\multicolumn{1}{l}{
\begin{tabular}{l}
Low Resolution / Infrared / Sketch / Text 
\end{tabular}} \\ \cline{2-3}

\multicolumn{1}{c}{} &
2021 IJCAI~\cite{lin2021survey2} & 
\multicolumn{1}{l}{
\begin{tabular}{l}
Deep Feature Representation Learning / Deep\\ Metric Learning / Identity-driven detection
\end{tabular}} \\ \cline{2-3}

\multicolumn{1}{c}{} &
2021 TPAMI~\cite{ye2021tpamisurvey} & 
\multicolumn{1}{l}{
\begin{tabular}{l}
Closed-world Setting (deep feature representation\\ learning; deep metric learning; ranking optimization)\\ / Open-world Setting (heterogeneous data; raw\\ images or videos; unavailable or limited labels;\\ open-set; noisy annotation)
\end{tabular}} \\ \cline{2-3}

\multicolumn{1}{c}{} &
2022 IVC~\cite{ming2022survey} & 
\multicolumn{1}{l}{
\begin{tabular}{l}
Deep Metric Learning / Local Feature Learning\\ / Generative Adversarial Networks / Sequence\\ Feature Learning / Graph Convolutional Networks
\end{tabular}} \\ \bottomrule
\end{tabular}}
\end{spacing}
\vspace{-0.45cm}
\end{table}

\section{Datasets and Evaluations}
\subsection{Datasets}
\label{datasets}
\begin{figure}  
\centering  
\includegraphics[width=0.50\textwidth]{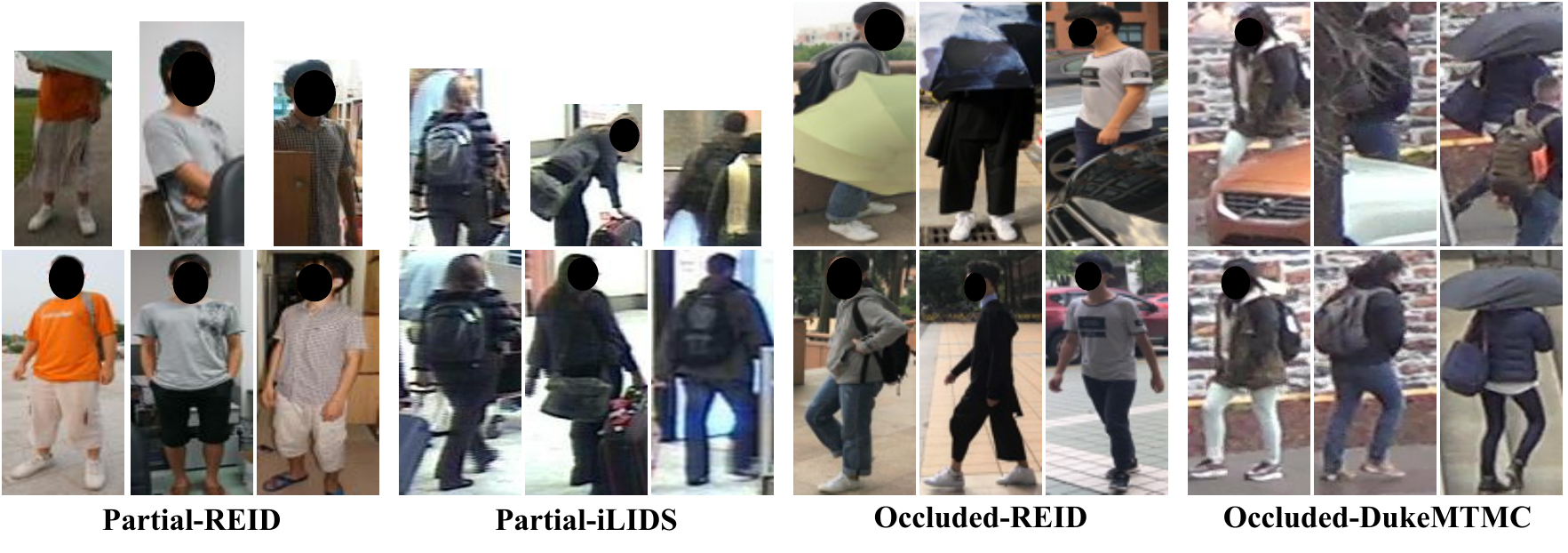}
\caption{Examples of partial/occluded person Re-ID datasets.
Partial/occluded person images (the first row) and full-body person images (the second row).}
\label{fig_dataset}
\end{figure}

We review eight widely-used datasets for occluded person Re-ID, including 3 image-based partial\footnote{The partial person Re-ID assumes that the visible region of occluded person image is manually cropped for identification.} Re-ID datasets, 4 image-based occluded\footnote{The occluded person Re-ID does not require the manually cropping process of occluded images.\\ Unless otherwise specified, occluded person Re-ID in this survey includes both partial person Re-ID and occluded person Re-ID.} Re-ID datasets, and 1 video-based occluded person Re-ID dataset.
The statistics of these datasets are summarized in Table~\ref{table_dataset} and the details of each dataset are reviewed as follows. 
Examples of partial/occluded person Re-ID datasets are shown in Fig.~\ref{fig_dataset}.

\noindent\textbf{Partial-REID}~\cite{zheng2015partial} is an image-based partial Re-ID dataset with a variety of viewpoints, backgrounds, and occlusion types.
It contains 600 images of 60 people, with 5 full-body images and 5 partial images per person.
The partial observation is generated by manually cropping the occluded region in occluded images.

\noindent\textbf{Partial-iLIDS}~\cite{he2018partial} is an image-based simulated partial Re-ID dataset derived from iLIDS~\cite{zheng2011iLIDS}.
%
% It has fair amount of occluded samples caused by other individuals or luggages.
%
It is captured by multiple non-overlapping cameras in the airport and contains 238 images from 119 people, with 1 full-body image and 1 manually cropped non-occluded partial image per person.

\noindent\textbf{p-CUHK03}~\cite{kim2017partial} is an image-based partial Re-ID dataset constructed from CUHK03~\cite{li2014deepreid}.
It contains 1360 person identities captured in campus environment.
In general, 1160 person identities are used as training set, 100 person identities are used as validation set, and 100 person identities are used as test set.
It selects 5 images with same view point from the raw dataset for each identity and generates 10 partial body probe images out of selected two images.
The rest 3 images of each identity are used as full body gallery image.

\noindent\textbf{P-ETHZ}~\cite{zhuo2018occluded} is an image-based occluded person Re-ID dataset modified from ETHZ~\cite{ess2008mobile}.
It has 3897 images of 85 person identities.
Each identity has 1 to 30 full-body person images and occluded person images respectively.

\noindent\textbf{Occluded-REID}~\cite{zhuo2018occluded} is an image-based occluded person Re-ID dataset captured by mobile cameras with different viewpoints and different types of severe occlusion.
It consists of 2000 images of 200 people, with 5 full-body images and 5 occluded images per person.
\begin{table}[!t]
\caption{Occluded/Partial Person Re-id Datasets. Occluded-dukemtmc and Occluded-dukemtmc-videoreid are Abbreviated as Occ-dukemtmc and Occ-dukemtmc-video Respectively.\label{table_dataset}}
\centering
\setlength \tabcolsep{1.3pt}
\begin{spacing}{1.15}
\resizebox{0.72\linewidth}{!}{%
\begin{tabular}{cccc}
\toprule
\multirow{2}{*}{Dataset} & \multirow{2}{*}{\begin{tabular}[c]{@{}c@{}}Training Set\\ (ID/Image)\end{tabular}} & \multicolumn{2}{c}{Test Set (ID/Image)} \\ \cline{2-4}
 &  & Gallery & Query \\ \cline{1-4}
Partial-REID & - & 60/300 & 60/300 \\
Partial-iLIDS & - & 119/119 & 119/119 \\
p-CUHK03 & 1160/15080 & 100/300 & 100/1000 \\ \cline{1-4}
P-ETHZ & 43/ - & 42/ - & 42/ - \\
Occluded-REID & - & 200/1000 & 200/1000 \\
P-DukeMTMC-reID & 650/ - & 649/ - & 649/ - \\
Occ-DukeMTMC & 702/15,618 & 1,110/17,661 & 519/2,210 \\ \cline{1-4}
Occ-DukeMTMC-Video & 702/292,343 & 1,110/281,114 & 661/39,526 \\ \bottomrule
\end{tabular}}
\end{spacing}
\vspace{-0.45cm}
\end{table}

\noindent\textbf{P-DukeMTMC-reID}~\cite{zhuo2018occluded} is an image-based occluded person Re-ID dataset modified from DukeMTMC-reID~\cite{zheng2017dukeimage}.
It has 24143 images of 1299 person identities and contains images with target persons occluded by different types of occlusion in public, \eg people, cars, and guideboards.
Each identity has both full-body images and occluded images.

\noindent\textbf{Occluded-DukeMTMC}~\cite{miao2019occluded} is an image-based occluded person Re-ID dataset built from DukeMTMC-reID~\cite{zheng2017dukeimage}.
It contains 15,618 images of 708 people for training while including 2,210 query images of 519 persons and 17,661 gallery images of 1110 persons for testing.
The 9\% of the training set, the 100\% of the query set, and the 10\% of the gallery set are occluded images.

\noindent\textbf{Occluded-DukeMTMC-VideoReID}~\cite{hou2021completion} is a video-based occluded Re-ID dataset reorganized from the DukeMTMC-VideoReID~\cite{wu2018dukevideo}.
It includes large variety of obstacles, \eg cars, trees, bicycles, and other persons.
It contains 1,702 image sequences of 702 identities for training, 661 query image sequences of 661 identities and 2,636 gallery image sequences of 1110 identities for testing.
More than 1/3 frames of each query sequence in the testing set contain occlusion.

\subsection{Evaluation Metrics}
The occluded person Re-ID evaluates the performance of a Re-ID system under the scenario of occlusion.
Therefore, the settings of partial/occluded person Re-ID datasets are usually specially designed.
In principle, the query images/videos for testing are all occluded samples.
The evaluation focuses on whether the correct identities can be retrieved when only occluded queries are provided.
To evaluate a Re-ID system, the Cumulative Matching Characteristics (CMC) curves and the mean Average Precision (mAP) are two widely used metrics.
The CMC curves calculate the probability that a correct match appears in the top-$k$ ranked retrieval results, $k \in \left\{ 1,2,3,...\right\}$.
Specifically, the top-$k$ accuracy of the query $i$ is calculated as:
\begin{small}
\begin{equation}
{Acc_k^i}=\begin{cases}
1 ,&{\text{if the top-$k$ ranked gallery samples}}\\
&{\text{contain the sample(s) of query $i$}};\\
0 ,&{\text{otherwise}}.
\end{cases}
\end{equation}
\end{small}
Supposing there are $N$ queries in the test set, the CMC-$k$ (\aka the rank-$k$ accuracy) that calculates the probability of the top-$k$ accuracy for all queries is computed as:
\begin{small}
\begin{equation}
\text{CMC-}k=\frac{1}{N}\sum_{i=1}^{N}Acc_k^i
\end{equation}
\end{small}
Since only the first match is concerned in the calculation, the CMC curves are acceptable when there are only one ground truth for each query or when we care more about the ground truth match in the top positions of the rank list.
The mAP measures the average retrieval performance that takes the order of all true matches in the ranked retrieval results into consideration.
Specifically, the average precision (AP) of the query $i$ is calculated as:
\begin{small}
\begin{equation}
\text{AP}_i=\frac{1}{M_i}\sum_{j=1}^{M_i}\frac{j}{Rank_j}
\end{equation}
\end{small}
where $M_i$ denotes there are $M_i$ samples with identity $i$ in the gallery set and $Rank_j$ denotes the rank of the $j$-th ground truth in the retrieval gallery list for query $i$.
Supposing there are $N$ queries in the test set, the mAP is computed as:
\begin{small}
\begin{equation}
\text{mAP}=\frac{1}{N}\sum_{i=1}^{N}\text{AP}_i
\end{equation}
\end{small}
Since the order of all true matches in the ranked retrieval results participates in the calculation of mAP, the mAP measures the average retrieval performance and is suitable for the gallery with multiple true matches.

On the whole, the mAP pays more attention to the ability of retrieval recall while the CMC curves focus on the ability of retrieving a true match in candidate lists with different sizes.
Consequently, the CMC curves and the mAP always work together for the evaluation of a Re-ID system.
\begin{figure}  
\centering  
\includegraphics[width=0.42\textwidth]{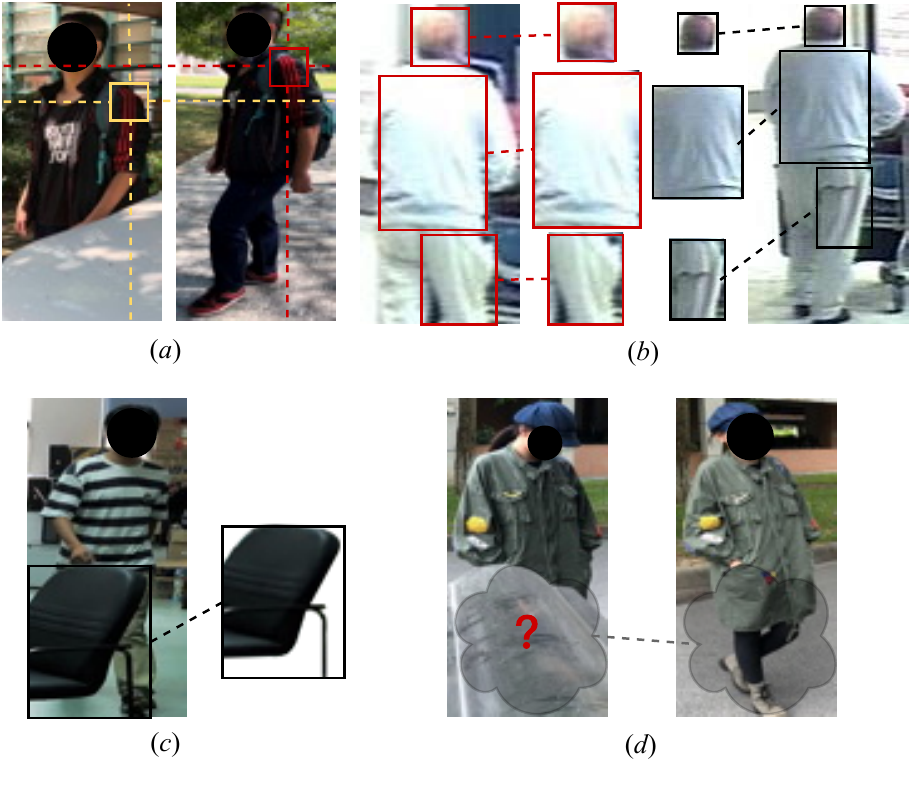}
\vspace{-0.3cm}
\caption{Examples of the four issues caused by occlusion in real-world applications: (\emph{a}) position misalignment, (\emph{b}) scale misalignment, (\emph{c}) noisy information, and (\emph{d}) missing information.}
\label{fig_four_issues_real}
\end{figure}

\section{Occluded Person Re-ID}
Occluded person re-identification (Re-ID) aims at addressing the occlusion problem when retrieving the person of interest across multiple cameras.
In real-world applications of person Re-ID, a person may be occluded by a variety of obstacles (\eg cars, trees, streetlights, and other persons) and the surveillance system often fails to capture the holistic person.
A practical person Re-ID system in video surveillance generally includes three stages: pedestrian detection, trajectory tracking, and person retrieval.
The occlusion will affect the whole process and bring great challenges to the final Re-ID.

In general, there are four issues to be considered when developing a solution for occluded person Re-ID.
The first two issues are the position and the scale misalignments between partial and holistic images.
This is mainly caused by the upstream data processing: the detected box of a partial person undergoes the same alignment processing as that of a holistic person to obtain a consistent input size, resulting in the position misalignment issue and the scale misalignment issue.
The last two issues are the noisy information and the missing information caused by occlusion.
In the detected boxes of occluded pedestrians, occlusion is inevitably included in whole or in part and the identity information of occluded regions is missing, resulting in the noisy information issue and the missing information issue.
Each issue is shown in Fig.~\ref{fig_four_issues} and real-world examples of the four issues are presented in Fig.~\ref{fig_four_issues_real}.

In this section, we analyze occlusion-related person Re-ID methods from the perspective of above-mentioned four issues.
Corresponding solutions for each issue are summarized following the taxonomy illustrated in Fig.~\ref{fig_taxonomy}.
It should be noticed that some methods simultaneously address more than one issue and these methods will be introduced multiple times from different perspectives accordingly.
%

% %
% \begin{figure*}  
% \centering  
% \includegraphics[width=0.86\textwidth]{fig_matching_attribute.pdf}
% \caption{(\emph{a}) The schematic diagram of matching-based methods: constructing multi-scale local matching elements and designing matching strategies to address the position misalignment, scale misalignment, and noise information issues.
% %
% (\emph{b}) The schematic diagram of attribute-based methods:  associating the semantic-level attribute annotations with person re-identification.
% }
% \label{fig_matching_attribute}
% \end{figure*}
% %

\subsection{\textbf{Position Misalignment}}
Deep learning-based solutions to address the position misalignment issue caused by occlusion in person Re-ID can be roughly summarized into four categories: matching~\cite{zheng2015partial,xu2021poseguide,he2018partial,he2020crowd,lin2021partial,jin2022occluded,wang2020occluded,yan2020videoreid,jia2021occluded}, auxiliary model for position~\cite{miao2019occluded,zhou2020occluded,miao2021identifying,han2020partial,gao2020occluded,zhai2021occluded,wang2020occluded,liu2021videoreid,he2021seriouslyMisalign,wang2022pfd,kalayeh2018parsingreid,quispe2019parsingreid,gao2020partial}, additional supervision for position~\cite{xu2018poseguide,zhang2019poseguide,xu2021poseguide,zheng2021poseguide,hou2021completion,cai2019multi,huang2020human,zhou2020fine}, and attention mechanism for position~\cite{kim2017partial,zhu2020identity,sun2019partial,huo2021partial,li2018diversity,wang2021occluded,li2021diverse,tan2022mhsa}.
Firstly, the matching-based solutions take person Re-ID as a matching task and propose a variety of matching components, as well as the matching strategies, to address the position misalignment issue.
Secondly, the auxiliary model-based solutions for person Re-ID rely on the position information provided by external models to boost performance. 
Thirdly, the additional supervision-based solutions for person Re-ID utilize extra information to guide the position-related learning process while being independent during the inference stage.
Fourthly, the attention mechanism-based solutions for person Re-ID learn attention to address the position misalignment issue without any additional information.

\subsubsection{\textbf{Matching}}
\begin{figure}  
\centering  
\includegraphics[width=0.4\textwidth]{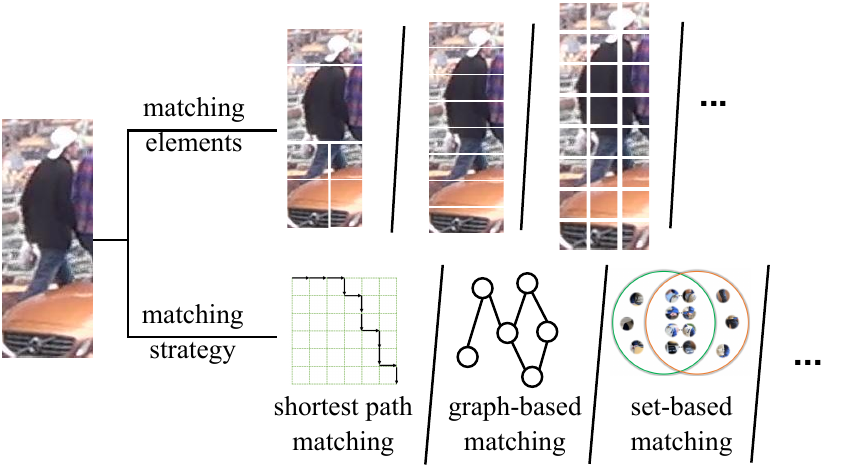}
\caption{The diagram of matching-based methods: constructing local matching elements and designing matching strategies to address the position misalignment, scale misalignment, and noise information issues.}\label{fig_method_matching}
\end{figure}

The main points of a matching-based method can probably be summarized as the matching component and the matching strategy (see Fig.~\ref{fig_method_matching}).
Diverse definitions of the matching component, as well as the corresponding matching strategies, have been proposed for addressing the position misalignment issue. 
On the whole, matching-based methods can be further grouped into sliding window matching~\cite{zheng2015partial}, shortest path matching~\cite{xu2021poseguide}, reconstruction-based matching~\cite{zheng2015partial,he2018partial}, denoising matching~\cite{he2020crowd,lin2021partial,jin2022occluded}, graph-based matching~\cite{wang2020occluded,yan2020videoreid}, and set-based matching~\cite{jia2021occluded}.

The sliding window matching~\cite{zheng2015partial} treats the partial probe image as a whole and slides it exhaustively over a full-body gallery image to match the most similar local region.
The L1-norm distance between the partial image and its most similar match on a full-body image is employed for the measurement.

The shortest path matching~\cite{xu2021poseguide} performs the matching by calculating the shortest path between two sets of local features and uses the matched local features to compute the similarity, explicitly accomplishing the position alignment.

The reconstruction-based matching~\cite{zheng2015partial,he2018partial} assumes the identity information in an occluded image is a subset of that in a non-occluded image and thus the partial image can be reconstructed in whole or in part from a holistic image of the same identity. 
The patch-level reconstruction-based matching~\cite{zheng2015partial} decomposes the partial and the full-body images into regular grid patches as matching components.
Each probe patch is matched to a set of gallery patches by optimizing the gallery patch selection for self reconstruction.
%
% The reconstruction residuals of each probe patch based on patches of a different person in the gallery can then be calculated and compared to determine the identity.
%
The block-level reconstruction-based matching~\cite{he2018partial} defines $c \times c$ pixels on a feature map as an independent block for matching.
It is assumed that each block of a partial probe image can be reconstructed from the sparse linear combination of blocks of a full-body gallery image with the same identity.

The denoising matching~\cite{he2020crowd,lin2021partial,jin2022occluded} proposes to focus on foreground visible human parts while discarding the noisy information during matching.
Specifically, GASM~\cite{he2020crowd} learns a saliency heatmap with the supervision of pose estimation and human segmentation to highlight foreground visible human parts.
Guided by the saliency heatmap, the matching scores of spatial element-wise features in foreground visible human parts are assigned with large weights while that in background or occlusion regions are assigned with small weights, adaptively.
Co-Attention~\cite{lin2021partial} performs the matching between a partial and a holistic image under the guidance of body parsing masks.
Particularly, the self-attention mechanism~\cite{vaswani2017attention} is adopted in Co-Attention matching where the parsing mask of the partial image is viewed as the query, parsing masks and CNN features of the partial and the holistic images serve as the key and the value respectively.
ASAN~\cite{jin2022occluded} proposes to replace the segmentation mask of holistic gallery images with the mask of the current target person image in each retrieval matching process.
In this way, the feature extraction guided by ASAN is able to suppress the interference from useless parts.

The graph-based matching~\cite{wang2020occluded,yan2020videoreid} formulates the occluded person re-identification as the graph matching problem.
HOReID~\cite{wang2020occluded} employs the key-point heatmaps to extract the semantic local features of an image as nodes of a graph.
The graph convolutional network (GCN) with learnable adjacent matrices is designed to pass messages between nodes for capturing the high-order relation information. 
To measure the similarity between two graphs, node features of the two graphs are further processed with relevant information extracted from each other for learning topology information.
Both the high-order relation information extracted by GCN and the topology information learned from each other are employed to compute the final similarity for the two graphs.
The multi-granular hypergraph matching~\cite{yan2020videoreid} designs multiple hypergraphs with different spatial and temporal granularities to address the misalignment and occlusion issues for video-based person re-identification.
Different from the conventional graphs, the hypergraphs~\cite{jiang2019hypergraph} are able to model the high-order dependency involving multiple nodes and are more suitable for modeling the multi-granular correlations in a sequence.

The set-based matching takes occluded person Re-ID as a set matching task without requiring explicit spatial alignment.
Specifically, MoS~\cite{jia2021occluded} employs a CNN backbone to capture diverse visual patterns along the channel dimension as matching elements.
And the Jaccard similarity coefficient is introduced as the metric to compute the similarity between pattern sets of person images.
The minimization and maximization are used to approximate the operations of intersection and union of sets, and the ratio of intersection over union is calculated to measure the similarity between two sets.
%
% Apart from this, the MoS also proposes a conflict penalty mechanism that detects mutually exclusive patterns in the pattern union of image pairs during inference and decreases their similarities accordingly to obtain better performance for matching.

% %
% \begin{figure*}  
% \centering  
% \includegraphics[width=0.95\textwidth]{fig_pose_seg.pdf}
% \caption{(\emph{a}) The schematic diagram of pose-based methods: the coordinates of keypoints for addressing the position misalignment and the prediction confidence for alleviating the noisy information.
% %
% and (\emph{b}) The schematic diagram of segmentation-based methods: part masks for position locating and noisy information excluding.}
% \label{fig_pose_seg}
% \end{figure*}
% %

\subsubsection{\textbf{Auxiliary Model for Position}}
To address the position misalignment issue caused by occlusion, some methods directly use the auxiliary information provided by external models for position alignment.
According to the type of employed auxiliary models, these methods can be roughly summarized into pose-based~\cite{miao2019occluded,zhou2020occluded,miao2021identifying,han2020partial,gao2020occluded,zhai2021occluded,wang2020occluded,liu2021videoreid,he2021seriouslyMisalign,wang2022pfd}, parsing-based~\cite{kalayeh2018parsingreid}, and hybrid-based~\cite{quispe2019parsingreid,gao2020partial} solutions.

The pose-based methods~\cite{miao2019occluded,zhou2020occluded,miao2021identifying,han2020partial,gao2020occluded,zhai2021occluded,wang2020occluded,liu2021videoreid,he2021seriouslyMisalign,wang2022pfd} utilize the position information predicted by an external pose estimation model to address the position misalignment issue.
PGFA~\cite{miao2019occluded}, PDVM~\cite{zhou2020occluded}, and PMFB~\cite{miao2021identifying} generate heatmaps consisting of a 2D Gaussian centered on key-point locations to extract aligned pose features through a dot product with the CNN feature map. 
KBFM~\cite{han2020partial} utilizes shared visible keypoints between images to locate aligned rectangular regions for calculating the similarity.
PVPM~\cite{gao2020occluded} uses the key-point heatmaps and the part affinity fields estimated by OpenPose~\cite{cao2017openpose} to generate part attention maps for extracting aligned local features.
Similarly, PGMANet~\cite{zhai2021occluded} generates heatmaps of key-points locations and employs them to calculate part attention masks on the CNN feature map.
Based on the part features aggregated by part attention masks, PGMANet further computes the correlation among different part features to exploit the second-order information to enrich the feature extraction.
HOReID~\cite{wang2020occluded} employs the key-point heatmaps to extract the semantic local features on a person image.
The local features of an image are taken as nodes of a graph and HOReID designs a graph convolutional network with learnable adjacent matrices to exploit the more discriminative relation information among nodes for re-identification.
Further, CTL~\cite{liu2021videoreid} proposes to divide the human body into three granularities and uses key-point heatmaps to extract multi-scale part features as graph nodes.
The cross-scale graph convolution and the 3D graph convolution are designed to capture the structural information and the hierarchical spatial-temporal dependencies for addressing the spatial misalignment issues in video person Re-ID.
ACSAP~\cite{he2021seriouslyMisalign} proposes to use the external pose information to guide the adversarial generation of aligned feature maps.
PFD~\cite{wang2022pfd} generates local patch features and multiplies them with the processed keypoint heatmaps element-wisely to obtain the pose-guided features.
Throught the measurement of set similarity, PFD performs the matching between local features and pose-guided features to disentangle the pose information from patch features for position alignment.
\begin{figure}  
\centering  
\includegraphics[width=0.44\textwidth]{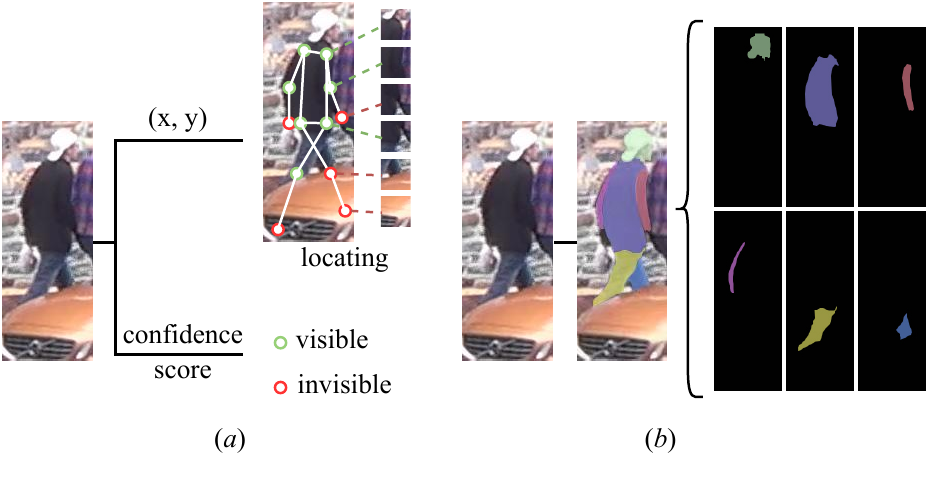}
\caption{(\emph{a}) The diagram of pose-based methods: keypoint coordinates for addressing the position misalignment and confidence scores for excluding the noisy information.
(\emph{b}) The diagram of segmentation-based methods: part masks for position locating and noisy information excluding.
}\label{fig_method_pose_seg}
\end{figure}

The parsing-based methods directly take advantage of the parsing information generated from an external human parsing model to address the position misalignment issue.
SPReID~\cite{kalayeh2018parsingreid} trains an extra semantic parsing model on the human parsing dataset LIP~\cite{gong2017parsing} to predict probability maps associated to 5 pre-defined semantic regions of human body.
These probability maps are then used to extract different semantic features through the weighted sum operation on the feature map generated by a modified Inception-V3~\cite{szegedy2016inceptionv3}.
% More solutions related to human parsing for addressing the position misalignment issue are categorized into \emph{Additional Supervision} since the parsing information are indirectly used, \ie employed for supervision, in these methods. 
% \textcolor{blue}{Although the SPReID is beneficial for addressing the position misalignment issue cause by occlusion, it should be noticed that the SPReID is proposed on the basis of the general person Re-ID task, and we summarize the SPReID here for a more comprehensive analysis of solutions related to occlusion (so as other general person Re-ID methods).}
% %
% Besides, more occluded Re-ID methods related to human parsing for addressing the position misalignment issue are categorized into \emph{Additional Supervision} since they only use the parsing information indirectly (employ the parsing information for supervision). 

The hybrid-based methods~\cite{quispe2019parsingreid,gao2020partial} employ two or more external models to provide auxiliary information for addressing the position misalignment issue.
Specifically, SSP-ReID~\cite{quispe2019parsingreid} exploits the capabilities of both clues, \ie the saliency and the semantic parsing, to guide the CNN backbone to learn complementary representations for re-identification.
The external off-the-shelf deep models~\cite{gong2017parsing,li2016deepsaliency} are employed to generate the semantic parsing masks and the saliency masks.
The element-wise product is applied between masks and a CNN feature map to obtain parsing features and saliency features for fusion.
TSA~\cite{gao2020partial} employs off-the-shelf HRNet~\cite{sun2019hrnet} and DensePose~\cite{guler2018densepose} to provide extra key-points information and body parts information.
Based on the key-points locations, the TSA divides the whole person into 5 regions and obtains region features on the CNN feature map through the soft region pooling.
Based on the body part masks, the TSA extracts corresponding region features on the texture image produced by a texture generator.
The region features guided by the key-points are then concatenated with the region features guided by the part masks accordingly to learn robust representations for re-identification.

\subsubsection{\textbf{Additional Supervision for Position}}
Different from the auxiliary model-based solutions that rely on extra information in both training and test phases, the additional supervision-based solutions for position misalignment only use the extra information to guide the learning process and are independent during inference.
According to the type of external information used, the additional supervision-based solutions for position misalignment can be coarsely summarized into pose-based~\cite{xu2018poseguide,zhang2019poseguide,xu2021poseguide,zheng2021poseguide,hou2021completion}, segmentation-based~\cite{cai2019multi,huang2020human}, and hybrid-based~\cite{zhou2020fine} methods.

The pose-based methods~\cite{xu2018poseguide,zhang2019poseguide,xu2021poseguide,zheng2021poseguide,hou2021completion} employ external pose information to guide the learning process for alleviating the position misalignment problem.
Specifically, AACN~\cite{xu2018poseguide} and DAReID~\cite{xu2021poseguide} learn part attention maps to locate and extract part features with the ground truth built from external pose information.
DSAG~\cite{zhang2019poseguide} and PGFL-KD~\cite{zheng2021poseguide} use features located by pose information to guide the feature learning process.
DSAG constructs a set of densely semantically aligned part images with the external pose information provided by DensePose~\cite{guler2018densepose}.
Taking the semantically aligned part images as the input, DSAG serves as a regulator to guide the feature learning on original images through the carefully designed fusion and loss.
Similarly, PGFL-KD uses external keypoint heatmaps to extract semantically aligned features.
The aligned features are then employed to regularize the global feature learning through knowledge distillation and interaction-based training.
Differently, the RFCNet~\cite{hou2021completion} trains an adaptive partition unit supervised by external pose information to split the CNN feature map into different regions and extract region features for further processing.

The segmentation-based methods~\cite{cai2019multi,huang2020human} utilize extra segmentation masks provided by the segmentation model, \eg human parsing model or scene segmentation model, to guide the learning process for addressing the position misalignment problem.
Specifically, MMGA~\cite{cai2019multi} learns to generate the whole-body, upper-body, and lower-body attention maps with the parsing labels estimated by JPPNet~\cite{liang2018jppnet}.
The learned attention maps are then employed to extract global and local features for re-identification.
HPNet~\cite{huang2020human} introduces human parsing as an auxiliary task and employs the parsing masks to extract part-level features for addressing the position misalignment issue.
The person Re-ID and the human parsing are learned in a multi-task manner where the pseudo parsing label are predicted by the scene segmentation model~\cite{fu2019danet} trained on the COCO DensePose~\cite{guler2018densepose} dataset.

The hybrid-based methods utilize more than one type of external information to guide the learning process.
Specifically, FGSA~\cite{zhou2020fine} mines fine-grained local features with the supervision of both the pose information and the attribute information to address the position misalignment issue.
FGSA designs a pose resolve net (pre-trained on MSCOCO~\cite{lin2014mscoco}) to provide part confidence maps and part affinity fields of the key parts.
These part maps are then used to extract part features on a CNN feature map through compact bilinear pooling.
Given the extracted part features, FGSA treats the attribute recognition as multiple classification tasks and trains an intermediate model for attribute classification along with the person Re-ID.
In this way, the attribute classification tasks guide the pose resolve net and the CNN backbone to learn more discriminated local information for re-identification.

\subsubsection{\textbf{Attention Mechanism for Position}}
\begin{figure}  
\centering  
\includegraphics[width=0.48\textwidth]{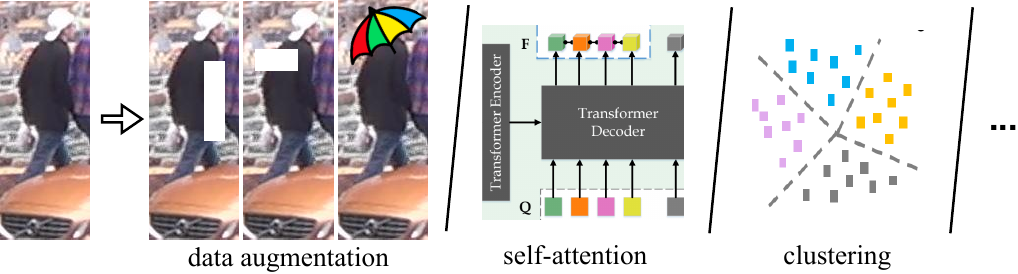}
\caption{Examples of technical routes in attention mechanism-based methods.
}\label{fig_method_attention}
\end{figure}

The attention mechanism-based methods learn attention to address the position misalignment issue without any additional information (see Fig.~\ref{fig_method_attention}).
According to the main point of the attention learning process, the attention mechanism-based solutions for position misalignment can be further grouped into cropping-based~\cite{kim2017partial}, clustering-based~\cite{zhu2020identity}, self-supervised~\cite{sun2019partial,huo2021partial}, and constraint-based~\cite{li2018diversity,wang2021occluded,li2021diverse,tan2022mhsa} methods.
It should be noticed that some methods have also involved the attention mechanism but rely on external information provided by auxiliary models or use additional information for supervision. 
These methods are summarized in \textit{Auxiliary Model for position} or \textit{Additional Supervision for Position} accordingly.

The cropping-based methods crop images into different local regions and learn attention to find the same local regions across different images for similarity calculation.
Specifically, DPPR~\cite{kim2017partial} crops 13 predefined partial regions on holistic images and designs an attention module conditioned on the partial probe image to assign the partial regions with larger attention weights if the same body parts are included.

The clustering-based methods generate pseudo-labels from clustering to supervise the attention learning.
Specifically, ISP~\cite{zhu2020identity} designs the cascaded clustering on CNN feature maps to gradually generate pixel-level pseudo-labels of human parts for part attention learning.
Based on the assumption that the foreground pixels have higher responses than the background ones, all pixels of a feature map are first clustered into foreground or background according to the activation.
Secondly, the foreground pixels are further clustered into different human parts according to the similarities between pixels.
The clustered pixel-level pseudo-labels of human parts are then employed to guide the part attention learning for extracting local part features to address the position misalignment issue.
%

% %
% \begin{figure*}  
% \centering  
% \includegraphics[width=0.95\textwidth]{fig_imgTrans_attention.pdf}
% \caption{(\emph{a}) The schematic diagram of image transformation methods: simultaneously addressing the position and the scale misalignment issues.
% %
% (\emph{b}) Examples of technical routes in attention mechanism methods for occluded person Re-ID.}
% \label{fig_imgTrans_attention}
% \end{figure*}
% %

The self-supervised methods~\cite{sun2019partial,huo2021partial} construct self-supervision to guide the attention learning.
Given a holistic image, VPM~\cite{sun2019partial} defines $m \times n$ rectangle regions and randomly crops a patch in which every pixel is assigned with the region label accordingly for self-supervision. 
VPM appends a region locator upon the extracted CNN feature map to discover different regions through the pixel-wise classification (attention) with the self-supervision constructed above.
APN~\cite{huo2021partial} randomly crops the holistic image into different partial types as the self-supervision for attention-based cropping type classification of a partial image.
The holistic gallery images are cropped for person retrieval according to the predicted cropping type of the partial probe image to accomplish the position alignment.

The constraint-based methods~\cite{li2018diversity,wang2021occluded,li2021diverse,tan2022mhsa} build constraints among different attention maps to help locate diverse body parts for addressing the position misalignment issue.
The multiple spatial attentions in~\cite{li2018diversity} employ a diversity regularization term on attention maps to ensure each attention focuses on different regions of the given image.
SBPA~\cite{wang2021occluded} separates local attention maps through minimizing the L1-norm distance between the local attention and the masked global attention.
PAT~\cite{li2021diverse} maintains vectors of part prototypes to generate part-aware attention masks on contextual CNN features and designs the part diversity mechanism to help achieve diverse part discovery.
Specifically, the part diversity mechanism minimizes the cosine similarity between every two part features to expand the discrepancy among different part features.
MHSA-Net~\cite{tan2022mhsa} proposes the feature diversity regularization term to encourage the diversity of local features captured by a multi-head self-attention mechanism.
Specifically, in order to obtain diverse local features, the regularization term restricts the Gram matrix of local features to be close to an identity matrix.

\subsection{\textbf{Scale Misalignment}}
Deep learning-based methods~\cite{zheng2019pyramidal,yan2020videoreid,he2018partial,he2019pyramid,wang2021occluded} propose to construct pyramid features or multi-scale features for alleviating the scale misalignment issue in person Re-ID.
Given a CNN feature map, the pyramid features are extracted from global to local while the multi-scale features maintain features of different receptive fields at the same position.
On the whole, the core of both pyramid features and multi-scale features is to extract features of different scales to construct robust representations to scale variation.

\subsubsection{\textbf{Pyramid Features}}
The pyramid features~\cite{zheng2019pyramidal,yan2020videoreid} are hierarchical and are extracted from global to local on the feature map.
The pyramidal model in~\cite{zheng2019pyramidal} horizontally slices the feature map into $n$ basic parts and builds corresponding branches for every $l \in \{1,2,...,n\}$ adjacent parts to obtain the pyramid features.
MGH~\cite{yan2020videoreid} hierarchically divides the feature map into $p \in \{1,2,4,8\}$ horizontal strips and average pools each strip to obtain multi-granular spatial features.

\subsubsection{\textbf{Multi-scale Features}}
The multi-scale features~\cite{he2018partial,he2019pyramid,wang2021occluded} focus on restricted local regions and maintain features of different receptive fields at the same position.
Specifically, DSR~\cite{he2018partial} average pools the square area of $s \times s$ pixels on a feature map to obtain the multi-scale block representations for alleviating the influence of scale mismatching, $s=\{1,2,3\}$. 
FPR~\cite{he2019pyramid} performs multiple max-pooling layers of different kernel sizes upon the feature map to capture diverse spatial features from small local regions to relatively large regions.
SBPA~\cite{wang2021occluded} maintains features in different scales at each pixel through the scale-wise residual connection for addressing the scale misalignment issue.

\subsection{\textbf{Position and Scale Misalignment}}
\begin{figure}  
\centering  
\includegraphics[width=0.48\textwidth]{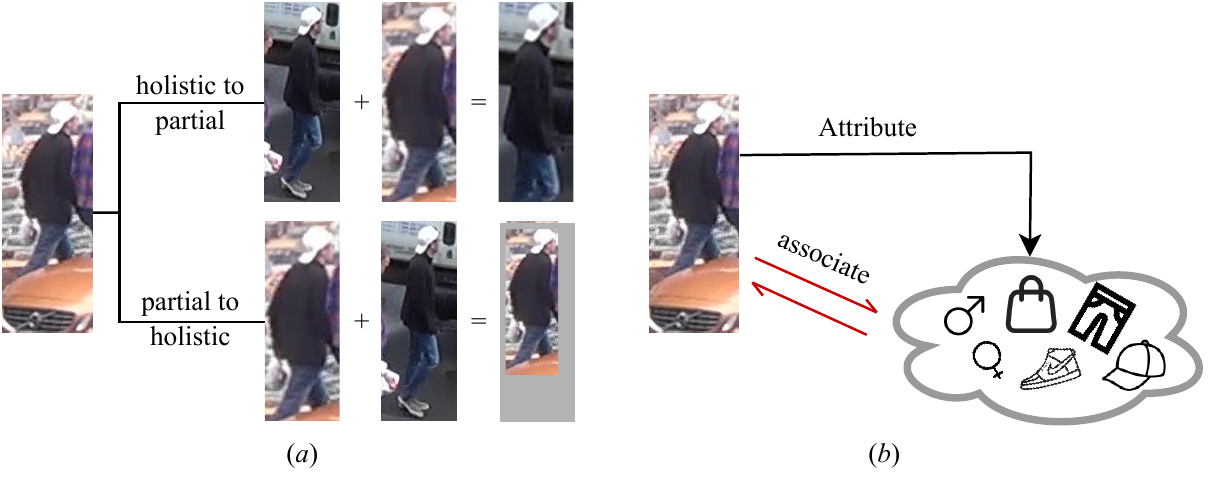}
\caption{(\emph{a}) The diagram of image transformation methods: addressing the position and scale misalignments simultaneously.
(\emph{b}) The diagram of attribute-based methods:  associating the attribute annotations with person Re-ID.
}\label{fig_method_imgTrans_attribute}
\end{figure}

In the literature, there are some methods~\cite{zhong2020robust,he2021partial,lin2021partial,he2021seriouslyMisalign,luo2020partial,huo2021partial} that simultaneously address the position and scale misalignment issues through the transformation of partial or holistic images (see Fig.~\ref{fig_method_imgTrans_attribute} (\emph{a})).
These methods are specifically summarized in this subsection for clarity.

The partial image transformation~\cite{zhong2020robust,he2021partial,lin2021partial,he2021seriouslyMisalign} aims to transform partial images to obtain the rectified results that are spatially aligned with holistic ones for re-identification.
Specifically, APNet~\cite{zhong2020robust} designs a bounding box aligner (BBA) which predicts 4 offset values (top, bottom, left, and right) to shift the detected bounding boxes to cover the estimated holistic body region.
Without manual annotations, APNet is trained by constructing automatic data augmentation.
PPCL~\cite{he2021partial} employs a gated transformation regression CNN module to predict the affine transformation coefficients between the partial and the holistic images for generating rectified partial images with proper scale and layout.
The prediction of transformation coefficients is self-supervised by randomly cropping holistic images to simulate partial images with known transformation coefficients.
The Image Rescaler (IR) in~\cite{lin2021partial} predicts the 2D affine transformation parameters to transform a partial image into a desirable distortion-free image for addressing the spatial misalignment issue.
Specifically, the partial image and the desirable distortion-free image are obtained by randomly cropping the holistic image and masking the uncropped regions, for self-supervision.
Differently, ACSAP~\cite{he2021seriouslyMisalign} designs a pose-guided generator that utilizes extra pose information of both partial and holistic images to guide the generation of aligned features for partial images.
The pose-guided generator is adversarially learned by training a pose-guided discriminator, which aims to distinguish the authenticity of image features.

The holistic image transformation~\cite{luo2020partial,huo2021partial} aims to find and transform the corresponding regions of holistic images to obtain the images that are spatially aligned with the query partial images for re-identification.
Specifically, STNReID~\cite{luo2020partial} utilizes the high-level CNN features of the partial and the holistic images to predict the 2D affine transformation parameters between them to transform the holistic image for matching with the partial image.
In STNReID, the holistic images are randomly cropped to partial images for building the self-supervised training of 2D affine transformation parameters prediction.
APN~\cite{huo2021partial} predicts the cropping type of a partial probe image and crops the corresponding regions of holistic gallery images for person retrieval.
In APN, the prediction of the cropping type is trained with the self-supervision constructed by randomly cropping the holistic image into different partial types.

\subsection{\textbf{Noisy Information}}
The occlusion is inevitably included in the detected boxes of occluded pedestrians and therefore brings noisy information to person Re-ID (see Fig.~\ref{fig_four_issues_real} (\emph{c})).
This is especially problematic since the similar obstacles across different identities disturb the appearance-based similarity calculation, \ie retrieval results easily contain negative images with similar obstacles~\cite{yu2021neighbourhood}.
Deep learning-based solutions for addressing the noisy information issue can be divided into auxiliary model for noise~\cite{miao2019occluded,zhou2020occluded,miao2021identifying,he2021seriouslyMisalign,han2020partial,gao2020occluded,yang2021learning,wang2022pfd,chen2019poseguide,gao2020partial,lin2021partial}, additional supervision for noise~\cite{xu2018poseguide,xu2021poseguide,he2019pyramid,huang2020human,zhang2021semantic,he2020crowd}, and attention mechanism for noise~\cite{zhuo2018occluded,sun2019partial,zhong2020random,zhong2020robust,zhao2021incremental,yan2021occluded,chen2021occlude,jia2022learning,wang2022fed,zheng2019re,zhao2020crowd,tan2021incomplete,kim2022ocnet}.

\subsubsection{\textbf{Auxiliary Model for Noise}}
The auxiliary model-based methods rely on external information provided by auxiliary models to help identify and suppress the noisy occlusion.
According to the type of employed auxiliary models, these methods can be further divided into pose-based~\cite{miao2019occluded,zhou2020occluded,miao2021identifying,he2021seriouslyMisalign,han2020partial,gao2020occluded,yang2021learning,wang2022pfd,chen2019poseguide} and parsing-based~\cite{gao2020partial,lin2021partial}.

The pose-based methods~\cite{miao2019occluded,zhou2020occluded,miao2021identifying,he2021seriouslyMisalign,han2020partial,gao2020occluded,yang2021learning,wang2022pfd,chen2019poseguide} exploits pose landmarks predicted by external pose estimation models to disentangle the useful information from the occlusion noise.
Specifically, PGFA~\cite{miao2019occluded}, PDVM~\cite{zhou2020occluded}, and PMFB~\cite{miao2021identifying} preset a threshold to filter out occluded invisible pose landmarks based on their prediction confidence.
The visible pose landmarks are used to generate heatmaps for extracting non-occluded pose features.
These methods also construct different local part features on the CNN feature map and employ the confidence of pose estimation to determine whether the corresponding parts are occluded or not.
Only the shared visible parts between probe and gallery images are used in these methods to compute the similarity, explicitly avoiding the disturbance from occlusion.
Similarly, ACSAP~\cite{he2021seriouslyMisalign} utilizes the confidence of pose estimation to decide the visibility of horizontally partitioned parts and assigns the shared visible parts with larger weights in similarity metrics.
KBFM~\cite{han2020partial} builds rectangular regions based on the shared visible keypoints between probe and gallery images to extract features for measuring the similarity.
Moreover, PMFB employs the pose embeddings generated from visible landmarks as gates to adaptively recalibrate channel-wise feature responses based on the visible body parts.
Given part features extracted with external pose information, PVPM~\cite{gao2020occluded} utilizes the characteristic of part correspondence between images of the same identity to mine correspondence scores as pseudo-labels for training a visibility predictor that estimates whether a part suffers from the occlusion.
Considering that the provided external pose information may be sparse or noisy, LKWS~\cite{yang2021learning} proposes to discretize the pose information to obtain robust visibility label of horizontally divided body parts.
Utilizing the label of visibility, LKWS trains a visibility discriminator to help suppress the influence of invisible horizontal parts.
Specifically, the visibility of each horizontal part is voted by all keypoints within the part region based on their prediction confidence scores and the preset threshold.
PFD~\cite{wang2022pfd} assigns the keypoint heatmap of higher confidence score than the preset threshold with $label=1$ and the keypoint heatmap of lower confidence score with $label=0$.
With the assigned labels, PFD divides the view feature set into high-confidence and low-confidence keypoint view feature sets.
The high-confidence keypoint view features are employed in the inference stage to explicitly match visible body parts and automatically separate occlusion features.
Moreover, PFD designs a Pose-guided push loss that encourages the difference between human body parts (\ie the high-confidence features) and non-human body parts (\ie the low-confidence features) to help focus on human body parts and alleviate the interference of occlusion.
Differently, STAL~\cite{chen2019poseguide} uses external pose landmarks to slice the video into multiple spatial-temporal units.
To down-weight the possible occluded units, STAL designs a joint spatial-temporal attention module to evaluate the quality scores of each unit with carefully designed loss functions.

The parsing-based methods~\cite{gao2020partial,lin2021partial} employ parsing masks estimated by human parsing models to help suppress the noisy occlusion.
Specifically, TSA~\cite{gao2020partial} utilizes external parsing results estimated by DensePose~\cite{guler2018densepose} to provide the visible signal to guide the learning of visible regions, suppressing the invisible occluded regions.
Co-Attention~\cite{lin2021partial} employs parsing masks as the query in the self-attention mechanism to perform image matching on associated regions for alleviating the noisy information brought by occlusion.

\subsubsection{\textbf{Additional Supervision for Noise}}
The additional supervision-based methods employ extra information to guide the learning of suppressing the occlusion noise while being independent during inference.
According to the type of the extra information employed, the additional supervision-based solutions can be further summarized into pose-based~\cite{xu2018poseguide,xu2021poseguide}, segmentation-based~\cite{he2019pyramid,huang2020human,zhang2021semantic}, attribute-based~\cite{jin2022occluded}, and hybrid-based~\cite{he2020crowd} methods.

The pose-based methods~\cite{xu2018poseguide,xu2021poseguide} utilize additional pose information to supervise the learning of excluding noisy occlusion.
Specifically, AACN~\cite{xu2018poseguide} uses external pose information to supervise the part attention learning.
Based on the intensities of each part attention map, AACN computes part visibility scores to measure the occlusion extent of each body part.
Similarly, with the ground truth built from external pose information, DAReID~\cite{xu2021poseguide} learns to predict upper and lower body masks to extract non-occluded part features.
Moreover, DAReID regards the heatmap of upper or lower features with large high-activation areas as reliable.
The significance of the reliable regions is enhanced to suppress the occlusion noise.

The segmentation-based methods~\cite{he2019pyramid,huang2020human,zhang2021semantic} utilize extra segmentation masks provided by a segmentation model, \eg the human parsing model or the scene segmentation model, to help address the noisy information issue.
Specifically, FPR~\cite{he2019pyramid} employs the person mask obtained by CE2P~\cite{ruan2019ce2p} to supervise the generation of foreground probability maps, encouraging the feature extraction to concentrate more on clean human body parts to refine the similarity computation with less contamination from occlusion.
HPNet~\cite{huang2020human} uses part labels provided by a COCO-trained human parsing model to learn human parsing and person re-identification in a multi-task manner. 
In HPNet, the predicted part probability maps are binarized with a threshold of 0.5 to extract part-level features and determine the visibility of each part for alleviating the occlusion noise.
SORN~\cite{zhang2021semantic} trains a semantic branch with pseudo-labels predicted by external semantic segmentation model to generate foreground-background masks for extracting features from non-occluded areas.

The attribute-based methods leverage semantic-level attribute annotations of person Re-ID datasets to help suppress the noisy information brought by occlusion (see Fig.~\ref{fig_method_imgTrans_attribute} (\emph{b})).
ASAN~\cite{jin2022occluded} employs the attribute information to guide the learning of occlusion-sensitive segmentation in a weakly supervised manner to extract non-occluded human body features.

The hybrid-based methods employ more than one type of external information to guide the learning process for alleviating the noisy information issue.
Specifically, GASM~\cite{he2020crowd} trains a mask layer and a pose layer with the ground truth predicted by the semantic segmentation model PSPNet~\cite{zhao2017parsing} and the pose estimation model CenterNet~\cite{zhou2019pose}.
GASM then combines the mask heatmap predicted by the mask layer and the keypoint heatmaps estimated by the pose layer into a saliency map to extract salient features, explicitly excluding the noisy information brought by occlusion.

\subsubsection{\textbf{Attention Mechanism for Noise}}
The attention mechanism-based methods learn to generate the attention maps that assign smaller weights to occluded regions to address the noisy information issue, without the requirement of any extra information.
According to the main idea of the attention learning process, the attention mechanism-based solutions for noise can be further grouped into data augmentation~\cite{zhuo2018occluded,sun2019partial,zhong2020random,zhong2020robust,zhao2021incremental,yan2021occluded,chen2021occlude,jia2022learning,wang2022fed}, query-guided~\cite{zheng2019re,zhao2020crowd}, drop-based~\cite{tan2021incomplete}, and relation-based~\cite{kim2022ocnet} methods.

The data augmentation methods~\cite{zhuo2018occluded,sun2019partial,zhong2020random,zhong2020robust,zhao2021incremental,yan2021occluded,chen2021occlude,jia2022learning,wang2022fed} generate artificial occlusion to train the network to focus on clean body parts and exclude the noisy occlusion.
Specifically, AFPB~\cite{zhuo2018occluded} constructs an occlusion simulator where a random patch from the background of source images is used as an occlusion to cover a part of holistic persons.
AFPB designs the multi-task losses that force the network to simultaneously identify the person and classify whether the sample is from the occluded data distribution.
VPM~\cite{sun2019partial} pre-defines $m \times n$ rectangle regions on an image and randomly crops partial pedestrian images from the holistic ones where every pixel in partial images is assigned with the region label accordingly for self-supervision.
In VPM, a region locator is designed to generate probability maps that infer the location of each region.
Through the sum operation over each probability map, VPM obtains the region visibility scores to suppress the occluded noisy regions.
RE~\cite{zhong2020random} proposes the operation of random erasing, which randomly selects a rectangle region in an image and erases its pixels with random values, to generate images with various levels of occlusion for training the model to extract non-occluded discriminative identity information.
APNet~\cite{zhong2020robust} trains a part identifier with the self-supervision built from data augmentation to identify visible strip parts.
The visible part features are then selected for similarity computation while the invisible part features on occluded or noisy regions are discarded.
IGOAS~\cite{zhao2021incremental} designs an incremental generative occlusion block that randomly generates a uniform occlusion mask from small to large on images in a batch, training the model more robust to occlusion through gradually learning harder occlusion.
With the synthesized occlusion data and their corresponding occlusion masks, IGOAS focuses more on foreground information by suppressing the response of generated occlusion regions to zero.
SSGR~\cite{yan2021occluded} employs the random erasing and the batch-constant erasing, which equally divides images into horizontal strips and randomly erases the same strip in a sub-batch, to simulate occlusion for training the disentangled non-local (DNL~\cite{yin2020dnl}) attention network.
OAMN~\cite{chen2021occlude} designs a novel occlusion augmentation scheme that crops a rectangular patch of a randomly chosen training image and scales the patch onto four pre-defined locations of the target image, producing diverse and precisely labeled occlusion.
With the supervision of labeled occlusion data, the OAMN learns to generate spatial attention maps which precisely capture body parts regardless of the occlusion.
DRL-Net~\cite{jia2022learning} utilizes the obstacles appearing in the train set to synthesize more diverse and realistic occluded samples to guide the contrast feature learning for mitigating the interference of occlusion noises.
Apart from this, DRL-Net designs a transformer that simultaneously maintains the ID-relevant and ID-irrelevant queries for disentangled representation learning.
Differently, FED~\cite{wang2022fed} considers the occlusion from not only the non-pedestrian obstacles but also the non-target pedestrians for augmentation.
To simulate reasonable non-pedestrian occlusions, FED manually crops the patches of backgrounds and occlusion objects from training images, and pastes them on pedestrian images with carefully designed augmentation process.
To synthesize non-target pedestrian occlusions, FED maintains a memory bank of the feature centers of different identities.
The memory bank is then employed to search $K$-nearest features of different identities for the current pedestrian.
Finally, FED diffuses the pedestrian representations with these memorized features and generates non-target pedestrian occlusions at the feature level for training.

The query-guided methods~\cite{zheng2019re,zhao2020crowd} aims to generate consistent attentions between query and gallery images for addressing the noisy information issue.
%query-gallery consistent attention
Specifically, CASN~\cite{zheng2019re} designs an attention-driven siamese learning architecture which enforces the attention consistency among images of the same identity to deal with viewpoint variations, occlusion, and background clutter.
%query-guided attention
PISNet~\cite{zhao2020crowd} focuses on the crowded occlusion in which the detected bounding boxes may involve multiple people and include distractive information.
Under the guidance of the query image (single-person), PISNet calculates the inner product of the query and gallery features to formulate the pixel-wise query-guided attention to enhance the feature of the target in the gallery image (multi-person).

The drop-based methods develop the training strategy based on dropping to guide the network to learn a more robust representation, alleviating the noisy information brought by occlusion.
%drop-based attention
Specifically, CBDB-Net~\cite{tan2021incomplete} uniformly partitions the feature map into strips and continuously drop each strip from top to bottom.
Trained with drop-based incomplete features, the model is forced to learn a more robust person descriptor for re-identification.

The relation-based methods mine the relation among different regions to refine features, alleviating the interference of occlusion.
Specifically, OCNet~\cite{kim2022ocnet} predefines the global region, top region (\ie $1/2$ top horizontal strip), bottom region (\ie $1/2$ bottom horizontal strip), and center region (\ie $1/3$ center vertical strip) on an image, and extracts four region features through the group convolution and the carefully designed attention mechanism.
In OCNet, the relational adaptive module consisting of two fully connected shared layers is proposed to capture the relation between different region features.
The relational weights are then used to refine region features to suppress the occluded or insignificant information.

\subsection{\textbf{Missing Information}}
The lack of identity information in occluded regions results in the missing information issue for person Re-ID (see Fig.~\ref{fig_four_issues} (\emph{d})).
There are mainly two ways to recover the missing information in occluded regions: spatial recovery~\cite{hou2019vrstc,hou2021completion} and temporal recovery~\cite{liu2019recovery,hou2019vrstc,hou2021completion}.

\subsubsection{\textbf{Spatial Recovery}}
The spatial recovery~\cite{hou2019vrstc,hou2021completion} utilizes the spatial structure of a pedestrian image to infer the missing information.
Specifically, VRSTC~\cite{hou2019vrstc} designs an auto-encoder which takes a image masked with white pixels as input and generates the contents for the occluded white region.
To improve the quality of generated contents of the occluded parts, VRSTC adopts a local and a global discriminator to adversarially judge the reality and the contextual consistency of the synthesized contents.
RFCNet~\cite{hou2021completion} exploits the long-range spatial contexts from non-occluded regions to predict the features of occluded regions, recovering the missing information at the feature level.
Specifically, RFCNet estimates four keypoints to divide the feature map into 6 regions.
In RFCNet, the encoder-decoder architecture is adopted, in which the encoder models the correlation between regions through clustering and the decoder utilizes the spatial correlation to recover occluded region features.

\subsubsection{\textbf{Temporal Recovery}}
The temporal recovery~\cite{liu2019recovery,hou2019vrstc,hou2021completion} requires the continuous sequence of images (\ie video-based person Re-ID) and utilizes the temporal information to recover the missing part in occluded regions.
Assuming that the information at the same position in other frames can help recover the lost information in the current frame, the Refining Recurrent Unit (RRU) in~\cite{liu2019recovery} is designed to remove noise and recover missing activation regions by implicitly referring the appearance and motion information extracted from historical frames.
VRSTC~\cite{hou2019vrstc} proposes a differentiable temporal attention layer which employs the cosine similarity to determine where to attend from adjacent frames for recovering the contents of the occluded parts.
RFCNet~\cite{hou2021completion} employs a query-memory attention mechanism in which the current region is regarded as the query and the corresponding regions of remaining frames serve as the memory.
The dot-product similarity~\cite{zhang2019self} between the query and each item of the memory is employed for re-weighting all items in the memory to obtain the long-term temporal contexts to refine the query.
\begin{table}[!t]
\centering
\newcommand{\bftab}[1]{{\fontseries{b}\selectfont#1}}
\caption{Performance Comparison on Partial-reid and Partial-ilids. S-ST is Short for Sing-ShoT, which Denotes that Each Identity Only Contains One Gallery Image in the Inference Stage.\label{table_performance_partial}}
\setlength \tabcolsep{1.4pt}
\begin{spacing}{1.06}
\resizebox{\linewidth}{!}{%
\begin{tabular}{ccccccccc}
\toprule
\multirow{2}{*}{Issues} & \multirow{2}{*}{\begin{tabular}[c]{@{}c@{}}Technical\\ Routes\end{tabular}} & \multicolumn{1}{c}{\multirow{2}{*}{Methods}} & \multicolumn{1}{c}{\multirow{2}{*}{Publications}} & \multicolumn{3}{c}{Partial-REID} & \multicolumn{2}{c}{Partial-iLIDS} \\ \cline{5-9}
 &  & \multicolumn{1}{c}{} & \multicolumn{1}{c}{} & Rank-1 & Rank-3 & S-ST & Rank-1 & Rank-3 \\ \cline{1-9}
\multicolumn{1}{c|}{\multirow{26}{*}{\begin{tabular}[c]{@{}c@{}}Position\\ Misalignment\end{tabular}}} & 

\multirow{6}{*}{Matching} & AMC+SWM~\cite{zheng2015partial} & ICCV2015 & 37.3 & 46.0 & \checkmark & 21.0 & 32.8 \\
\multicolumn{1}{c|}{} &  & DSR~\cite{he2018partial} & CVPR2018 & 50.7 & 70.0 & \checkmark & 58.8 & 67.2 \\
\multicolumn{1}{c|}{} &  & DAReID~\cite{xu2021poseguide} & KBS2021 & 68.1 & 79.5 & × & 76.7 & 85.3 \\
\multicolumn{1}{c|}{} &  & APN~\cite{huo2021partial} & ICPR2021 & 71.8 & 85.5 & × & 66.4 & 76.5 \\
\multicolumn{1}{c|}{} &  & Co-Attention~\cite{lin2021partial} & ICIP2021 & 83.0 & 90.3 & × & 73.1 & 83.2 \\
\multicolumn{1}{c|}{} &  & ASAN~\cite{jin2022occluded} & TCSVT2021 & \bftab{86.8} & \bftab{93.5} & × & \bftab{81.7} & \bftab{88.3} \\ \cline{2-9}

\multicolumn{1}{c|}{} & \multirow{8}{*}{\begin{tabular}[c]{@{}c@{}}Auxiliary Model\\ for Position\end{tabular}} & PDVM~\cite{zhou2020occluded} & PRL2020 & 43.3 & - & × & - & - \\
\multicolumn{1}{c|}{} &  & PGFA~\cite{miao2019occluded} & ICCV2019 & 68.0 & 80.0 & \checkmark & 69.1 & 80.9 \\
\multicolumn{1}{c|}{} &  & KBFM~\cite{han2020partial} & ICIP2020 & 69.7 & 82.2 & × & 64.1 & 73.9 \\
\multicolumn{1}{c|}{} &  & PMFB~\cite{miao2021identifying} & TNNLS2021 & 72.5 & 83.0 & × & 70.6 & 81.3 \\
\multicolumn{1}{c|}{} &  & TSA~\cite{gao2020partial} & ACM MM2020 & 72.7 & 85.2 & \checkmark & 73.9 & 84.7 \\
\multicolumn{1}{c|}{} &  & ACSAP~\cite{he2021seriouslyMisalign} & ICIP2021 & 77.0 & 83.7 & × & \bftab{76.5} & \bftab{87.4} \\
\multicolumn{1}{c|}{} &  & PVPM~\cite{gao2020occluded} & CVPR2020 & 78.3 & - & \checkmark & - & - \\
\multicolumn{1}{c|}{} &  & HOReID~\cite{wang2020occluded} & CVPR2020 & \bftab{85.3} & \bftab{91.0} & × & 72.6 & 86.4 \\ \cline{2-9}

\multicolumn{1}{c|}{} & \multirow{3}{*}{\begin{tabular}[c]{@{}c@{}}Additional\\ Supervision\\ for Position\end{tabular}} & DAReID~\cite{xu2021poseguide} & KBS2021 & 68.1 & 79.5 & × & \bftab{76.7} & 85.3 \\
\multicolumn{1}{c|}{} &  & PGFL-KD~\cite{zheng2021poseguide} & MM2021 & 85.1 & 90.8 & × & 74.0 & \bftab{86.7} \\
\multicolumn{1}{c|}{} &  & HPNet~\cite{huang2020human} & ICME2020 & \bftab{85.7} & - & \checkmark & 68.9 & 80.7 \\ \cline{2-9}

\multicolumn{1}{c|}{} & \multirow{4}{*}{\begin{tabular}[c]{@{}c@{}}Attention\\ Mechanism\\ for Position\end{tabular}} & VPM~\cite{sun2019partial} & CVPR2019 & 67.7 & 81.9 & \checkmark & 65.5 & 74.8 \\
\multicolumn{1}{c|}{} &  & APN~\cite{huo2021partial} & ICPR2021 & 71.8 & 85.5 & × & 66.4 & 76.5 \\
\multicolumn{1}{c|}{} &  & MHSA-Net~\cite{tan2022mhsa} & TNNLS2022 & 85.7 & 91.3 & × & 74.9 & 87.2 \\
\multicolumn{1}{c|}{} &  & PAT~\cite{li2021diverse} & CVPR2021 & \bftab{88.0} & \bftab{92.3} & × & \bftab{76.5} & \bftab{88.2} \\ \cline{2-9}
 
\multicolumn{1}{c|}{} & \multirow{5}{*}{\begin{tabular}[c]{@{}c@{}}Image\\ Transformation\end{tabular}} & STNReID~\cite{luo2020partial} & TMM2020 & 66.7 & 80.3 & × & 54.6 & 71.3 \\
\multicolumn{1}{c|}{} &  & APN~\cite{huo2021partial} & ICPR2021 & 71.8 & 85.5 & × & 66.4 & 76.5 \\
\multicolumn{1}{c|}{} &  & ACSAP~\cite{he2021seriouslyMisalign} & ICIP2021 & 77.0 & 83.7 & × & \bftab{76.5} & \bftab{87.4} \\
\multicolumn{1}{c|}{} &  & Co-Attention~\cite{lin2021partial} & ICIP2021 & 83.0 & \bftab{90.3} & × & 73.1 & 83.2 \\
\multicolumn{1}{c|}{} &  & PPCL~\cite{he2021partial} & CVPR2021 & \bftab{83.7} & 88.7 & × & 71.4 & 85.7 \\ \hline

\multicolumn{1}{c|}{\multirow{7}{*}{\begin{tabular}[c]{@{}c@{}}Scale\\ Misalignment\end{tabular}}} & \multirow{2}{*}{\begin{tabular}[c]{@{}c@{}}Multi-scale\\ Features\end{tabular}} & DSR~\cite{he2018partial} & CVPR2018 & 50.7 & 70.0 & \checkmark & 58.8 & 67.2 \\
\multicolumn{1}{c|}{} &  & FPR~\cite{he2019pyramid} & ICCV2019 & \bftab{81.0} & - & × & \bftab{68.1} & - \\ \cline{2-9}

\multicolumn{1}{c|}{} & \multirow{5}{*}{\begin{tabular}[c]{@{}c@{}}Image\\ Transformation\end{tabular}} & STNReID~\cite{luo2020partial} & TMM2020 & 66.7 & 80.3 & × & 54.6 & 71.3 \\
\multicolumn{1}{c|}{} &  & APN~\cite{huo2021partial} & ICPR2021 & 71.8 & 85.5 & × & 66.4 & 76.5 \\
\multicolumn{1}{c|}{} &  & ACSAP~\cite{he2021seriouslyMisalign} & ICIP2021 & 77.0 & 83.7 & × & \bftab{76.5} & \bftab{87.4} \\
\multicolumn{1}{c|}{} &  & Co-Attention~\cite{lin2021partial} & ICIP2021 & 83.0 & \bftab{90.3} & × & 73.1 & 83.2 \\
\multicolumn{1}{c|}{} &  & PPCL~\cite{he2021partial} & CVPR2021 & \bftab{83.7} & 88.7 & × & 71.4 & 85.7 \\ \hline

\multicolumn{1}{c|}{\multirow{16}{*}{\begin{tabular}[c]{@{}c@{}}Noisy\\ Information\end{tabular}}} & \multirow{9}{*}{\begin{tabular}[c]{@{}c@{}}Auxiliary Model\\ for Noise\end{tabular}} & PDVM~\cite{zhou2020occluded} & PRL2020 & 43.3 & - & × & - & - \\
\multicolumn{1}{c|}{} &  & PGFA~\cite{miao2019occluded} & ICCV2019 & 68.0 & 80.0 & \checkmark & 69.1 & 80.9 \\
\multicolumn{1}{c|}{} &  & KBFM~\cite{han2020partial} & ICIP2020 & 69.7 & 82.2 & × & 64.1 & 73.9 \\
\multicolumn{1}{c|}{} &  & PMFB~\cite{miao2021identifying} & TNNLS2021 & 72.5 & 83.0 & × & 70.6 & 81.3 \\
\multicolumn{1}{c|}{} &  & TSA~\cite{gao2020partial} & ACM MM2020 & 72.7 & 85.2 & \checkmark & 73.9 & 84.7 \\
\multicolumn{1}{c|}{} &  & ACSAP~\cite{he2021seriouslyMisalign} & ICIP2021 & 77.0 & 83.7 & × & 76.5 & 87.4 \\
\multicolumn{1}{c|}{} &  & PVPM~\cite{gao2020occluded} & CVPR2020 & 78.3 & - & \checkmark & - & - \\
\multicolumn{1}{c|}{} &  & Co-Attention~\cite{lin2021partial} & ICIP2021 & 83.0 & 90.3 & × & 73.1 & 83.2 \\
\multicolumn{1}{c|}{} &  & LKWS~\cite{yang2021learning} & ICCV2021 & \bftab{85.7} & \bftab{93.7} & × & \bftab{80.7} & \bftab{88.2} \\ \cline{2-9}

\multicolumn{1}{c|}{} & \multirow{3}{*}{\begin{tabular}[c]{@{}c@{}}Additional\\ Supervision\\ for Noise\end{tabular}} & DAReID~\cite{xu2021poseguide} & KBS2021 & 68.1 & 79.5 & × & 76.7 & 85.3 \\
\multicolumn{1}{c|}{} &  & FPR~\cite{he2019pyramid} & ICCV2019 & 81.0 & - & × & 68.1 & - \\
\multicolumn{1}{c|}{} &  & HPNet~\cite{huang2020human} & ICME2020 & \bftab{85.7} & - & \checkmark & \bftab{68.9} & \bftab{80.7} \\ \cline{2-9}

\multicolumn{1}{c|}{} & \multirow{4}{*}{\begin{tabular}[c]{@{}c@{}}Attention\\ Mechanism\\ for Noise\end{tabular}} & CBDB-Net~\cite{tan2021incomplete} & TCSVT2021 & 66.7 & 78.3 & × & 68.4 & 81.5 \\
\multicolumn{1}{c|}{} &  & VPM~\cite{sun2019partial} & CVPR2019 & 67.7 & 81.9 & \checkmark & 65.5 & 74.8 \\
\multicolumn{1}{c|}{} &  & FED~\cite{wang2022fed} & CVPR2022 & 83.1 & - & × & - & - \\
\multicolumn{1}{c|}{} &  & OAMN~\cite{chen2021occlude} & ICCV2021 & \bftab{86.0} & - & × & \bftab{77.3} & - \\ \bottomrule
\end{tabular}}
\end{spacing}
\vspace{-0.5cm}
\end{table}

\section{Discussion}
In this section, we summarize and analyze the evaluation results of occluded person Re-ID methods based on the four issues discussed earlier.
Following the proposed taxonomy, we aim to present potential factors that boost the performance of occluded person Re-ID to help facilitate future research.

\subsection{Performance Comparison}
\label{performance_comparison}
We evaluate the occluded person re-identification methods on four widely-used image-based datasets, \ie Partial-REID~\cite{zheng2015partial}, Partial-iLIDS~\cite{he2018partial}, Occluded-DukeMTMC (Occ-DukeMTMC)~\cite{miao2019occluded}, and Occluded-REID (Occ-ReID)~\cite{zhuo2018occluded}.
Details about the four datasets are illustrated in Sec.~\ref{datasets}.
Since the Partial-REID, Partial-iLIDS, and Occluded-REID are small, methods are generally trained on the training set of Market-1501~\cite{zheng2015market1501} and tested on these three datasets for evaluation.
The performance comparisons on two partial person Re-ID datasets and two occluded person Re-ID datasets are summarized in Table~\ref{table_performance_partial} and Table~\ref{table_performance_occluded} respectively.
From the two tables we obtain the following three observations:

1). For addressing the position misalignment issue and the noisy information issue, effective solutions are rich and diverse.
To be specific, firstly, there is not a dominant technical route to accomplish the position alignment:
On Partial-ReID, attention mechanism-based method PAT~\cite{li2021diverse} has achieved the top rank-1 accuracy; 
On Partial-iLIDS, matching-based method ASAN~\cite{jin2022occluded} has reached the best rank-1 and rank-3 accuracy; 
On Occluded-DukeMTMC, auxiliary model-based method PFD~\cite{wang2022pfd} has achieved the top rank-1 accuracy and the best mAP; 
On Occluded-ReID, additional supervision-based method HPNet~\cite{huang2020human} has obtained the best rank-1 accuracy.
Secondly, the most promising technical route to suppress the noisy occlusion is difficult to judge:
On Partial-ReID, the rank-1 accuracies of three technical routes, \ie auxiliary model, additional mechanism, and attention mechanism for noise, are comparable; 
On Partial-iLIDS and Occluded-DukeMTMC, auxiliary model-based methods LKWS~\cite{yang2021learning} and PFD~\cite{wang2022pfd} have reached the best rank-1 accuracy accordingly; 
On Occluded-ReID, additional supervision-based method HPNet~\cite{huang2020human} has achieved the top rank-1 accuracy.
Although neither the position misalignment issue nor the noisy information issue has a dominant technical route, the advantages and disadvantages of different technical routes can be summarized (see Sec.\ref{future_directions}).

2). The scale misalignment issue and the missing information issue have drawn less attention in existing methods.
Compared with the methods for addressing the position misalignment issue or the noisy information issue, the number of methods intended for the scale misalignment issue or the missing information issue is significantly smaller.

3). There are a large number of methods that have considered more than one issue at the same time.
For instance, PPCL~\cite{he2021partial} addresses the position and scale misalignment issues; PFD~\cite{wang2022pfd} focuses on the position misalignment and the noisy information issues; Co-Attention~\cite{lin2021partial} takes the position misalignment, the scale misalignment, and the noisy information issues into consideration.
However, there are none of the methods that have considered all the four issues caused by occlusion.
The in-depth analysis of issues and solutions in this survey aims to fill this gap and help develop a more comprehensive solution for occluded person Re-ID.

\subsection{Future Directions}
\label{future_directions}
As shown in Table~\ref{table_performance_partial} and Table~\ref{table_performance_occluded}, there have been consistent improvements in different technical routes for addressing different issues over the past few years.
Based on the analysis of issues and solutions, the following insights can be drawn for the future research of occluded person Re-ID.

From the perspective of four significant issues caused by occlusion, the position misalignment and the noisy information issues have been widely studied while the scale misalignment and the missing information issues are rarely considered in existing methods.
With the four issues summarized and analyzed in this survey, the deeper understanding of occluded person Re-ID can be obtained to contribute a more comprehensive solution and help inspire new ideas in the filed.

From the perspective of promising technical routes, it remains an open question since the evaluation results of state-of-the-art methods in different technical routes are comparable.
In spite of this, we analyze and summarize the advantages and disadcantages of different technical routes as follows to help boost future research.

1). \emph{Matching.}
The well-designed matching components and matching strategies greatly improves the performance of occluded person Re-ID.
Local and scalable matching components with the corresponding matching strategy can help address the position misalignment, scale misalignment, and noisy information issues.
Moreover, it can be easily integrated with other technical routes, \eg the repeatedly reported methods Co-Attention~\cite{lin2021partial} and ASAN~\cite{jin2022occluded} in Table~\ref{table_performance_partial}.

2). \emph{Auxiliary Model and Additional Supervision.}
In general, there are mainly three types of extra information employed for occluded person Re-ID: poses, segments, and attributes.
The position information, as well as their estimation confidence, provided by pose estimation or segmentation are used to help address the position misalignment and the noisy information issues respectively (see Fig.~\ref{fig_method_pose_seg}).
And the extra attribute information provided by person Re-ID datasets is generally utilized to formulate an extra task to help alleviate the issues brought by occlusion.
The two technical routes bring a lot of benefits and convenience while they are dependent on the extra labels or external models.

3). \emph{Attention Mechanism.}
The attention mechanism has been widely studied in existing methods for its huge potential and flexibility.
In recent three years, the methods~\cite{dosovitskiy2020vit,lin2021partial,jia2022learning,tan2022mhsa,wang2022fed} introduce the self-attention (Transformer) to occluded person Re-ID have made remarkable improvements on public datasets.
Similar to matching-based technical route, the attention mechanism can also be integrated with other technical routes, \eg APN~\cite{huo2021partial} in Table~\ref{table_performance_occluded}.

4). \emph{Image Transformation.}
The partial (holistic) image is transformed to obtain the image of consistent contents with the holistic (partial) image, addressing the position and the scale misalignment issues simultaneously (see Fig.~\ref{fig_method_imgTrans_attribute} (\emph{a})).
This technical route does make sense and is close to the ideal process while it requires more computation costs for the conditional image transformation in the inference stage.
Furthermore, the image transformation has not achieved a satisfying result in the current stage and the performance of this technical route is a little bit lower than others.
\begin{table}[!t]
\centering
\newcommand{\bftab}[1]{{\fontseries{b}\selectfont#1}}
\caption{Performance Comparison on Occ-dukemtmc and Occ-reid.\label{table_performance_occluded}}
\setlength \tabcolsep{1.4pt}
\begin{spacing}{1.06}
\resizebox{\linewidth}{!}{%
\begin{tabular}{cccccccc}
\toprule
\multirow{2}{*}{Issues} & \multirow{2}{*}{\begin{tabular}[c]{@{}c@{}}Technical\\ Routes\end{tabular}} & \multicolumn{1}{c}{\multirow{2}{*}{Methods}} & \multicolumn{1}{c}{\multirow{2}{*}{Publications}} & \multicolumn{2}{c}{Occ-DukeMTMC} & \multicolumn{2}{c}{Occ-ReID} \\ \cline{5-8}
 &  & \multicolumn{1}{c}{} & \multicolumn{1}{c}{} & Rank-1 & mAP & Rank-1 & mAP \\ \hline
\multicolumn{1}{c|}{\multirow{20}{*}{\begin{tabular}[c]{@{}c@{}}Position\\ Misalignment\end{tabular}}} & \multirow{6}{*}{Matching} & AMC+SWM~\cite{zheng2015partial} & ICCV2015 & - & - & 31.1 & 27.3 \\
\multicolumn{1}{c|}{} &  & GASM~\cite{he2020crowd} & ECCV2020 & - & - & 74.5 & 65.6 \\
\multicolumn{1}{c|}{} &  & DSR~\cite{he2018partial} & CVPR2018 & 40.8 & 30.4 & 72.8 & 62.8 \\
\multicolumn{1}{c|}{} &  & HOReID~\cite{wang2020occluded} & CVPR2020 & 55.1 & 43.8 & \bftab{80.3} & \bftab{70.2} \\
\multicolumn{1}{c|}{} &  & DAReID~\cite{xu2021poseguide} & KBS2021 & 63.4 & 53.2 & - & - \\
\multicolumn{1}{c|}{} &  & MoS~\cite{jia2021occluded} & AAAI2021 & \bftab{66.6} & \bftab{55.1} & - & - \\ \cline{2-8}

\multicolumn{1}{c|}{} & \multirow{6}{*}{\begin{tabular}[c]{@{}c@{}}Auxiliary Model\\ for Position\end{tabular}} & PVPM~\cite{gao2020occluded} & CVPR2020 & - & - & 70.4 & 61.2 \\
\multicolumn{1}{c|}{} &  & PGFA~\cite{miao2019occluded} & ICCV2019 & 51.4 & 37.3 & - & - \\
\multicolumn{1}{c|}{} &  & PDVM~\cite{zhou2020occluded} & PRL2020 & 53.0 & 38.1 & - & - \\
\multicolumn{1}{c|}{} &  & HOReID~\cite{wang2020occluded} & CVPR2020 & 55.1 & 43.8 & 80.3 & 70.2 \\
\multicolumn{1}{c|}{} &  & PMFB~\cite{miao2021identifying} & TNNLS2021 & 56.3 & 43.5 & - & - \\
\multicolumn{1}{c|}{} &  & PFD~\cite{wang2022pfd} & AAAI2022 & \bftab{69.5} & \bftab{61.8} & \bftab{81.5} & \bftab{83.0} \\ \cline{2-8}

\multicolumn{1}{c|}{} & \multirow{4}{*}{\begin{tabular}[c]{@{}c@{}}Additional\\ Supervision\\ for Position\end{tabular}} & HPNet~\cite{huang2020human} & ICME2020 & - & - & \bftab{87.3} & \bftab{77.4} \\
\multicolumn{1}{c|}{} &  & PGFL-KD~\cite{zheng2021poseguide} & MM2021 & 63.0 & 54.1 & 80.7 & 70.3 \\
\multicolumn{1}{c|}{} &  & DAReID~\cite{xu2021poseguide} & KBS2021 & 63.4 & 53.2 & - & - \\
\multicolumn{1}{c|}{} &  & RFCNet~\cite{hou2021completion} & TPAMI2021 & \bftab{63.9} & \bftab{54.5} & - & - \\ \cline{2-8}

\multicolumn{1}{c|}{} & \multirow{4}{*}{\begin{tabular}[c]{@{}c@{}}Attention\\ Mechanism\\ for Position\end{tabular}} &
MHSA-Net~\cite{tan2022mhsa} & TNNLS2022 & 59.7 & 44.8 & - & - \\
\multicolumn{1}{c|}{} &  & ISP~\cite{zhu2020identity} & ECCV2020 & 62.8 & 52.3 & - & - \\
\multicolumn{1}{c|}{} &  & PAT~\cite{li2021diverse} & CVPR2021 & \bftab{64.5} & 53.6 & \bftab{81.6} & \bftab{72.1} \\
\multicolumn{1}{c|}{} &  & SBPA~\cite{wang2021occluded} & SPL2021 & \bftab{64.5} & \bftab{54.0} & - & - \\ \hline

\multicolumn{1}{c|}{\multirow{2}{*}{\begin{tabular}[c]{@{}c@{}}Scale\\ Misalignment\end{tabular}}} & \multirow{2}{*}{\begin{tabular}[c]{@{}c@{}}Multi-scale\\ Features\end{tabular}} & FPR~\cite{he2019pyramid} & ICCV2019 & - & - & 78.3 & 68.0 \\
\multicolumn{1}{c|}{} &  & DSR~\cite{he2018partial} & CVPR2018 & 40.8 & 30.4 & 72.8 & 62.8 \\ \hline

\multicolumn{1}{c|}{\multirow{15}{*}{\begin{tabular}[c]{@{}c@{}}Noisy\\ Information\end{tabular}}} & \multirow{6}{*}{\begin{tabular}[c]{@{}c@{}}Auxiliary Model\\ for Noise\end{tabular}} & PVPM~\cite{gao2020occluded} & CVPR2020 & - & - & 70.4 & 61.2 \\
\multicolumn{1}{c|}{} &  & PGFA~\cite{miao2019occluded} & ICCV2019 & 51.4 & 37.3 & - & - \\
\multicolumn{1}{c|}{} &  & PDVM~\cite{zhou2020occluded} & PRL2020 & 53.0 & 38.1 & - & - \\
\multicolumn{1}{c|}{} &  & PMFB~\cite{miao2021identifying} & TNNLS2021 & 56.3 & 43.5 & - & - \\
\multicolumn{1}{c|}{} &  & LKWS~\cite{yang2021learning} & ICCV2021 & 62.2 & 46.3 & 81.0 & 71.0 \\
\multicolumn{1}{c|}{} &  & PFD~\cite{wang2022pfd} & AAAI2022 & \bftab{69.5} & \bftab{61.8} & \bftab{81.5} & \bftab{83.0} \\
\cline{2-8}

\multicolumn{1}{c|}{} & \multirow{4}{*}{\begin{tabular}[c]{@{}c@{}}Additional\\ Supervision\\ for Noise\end{tabular}} & GASM~\cite{he2020crowd} & ECCV2020 & - & - & 74.5 & 65.6 \\
\multicolumn{1}{c|}{} &  & FPR~\cite{he2019pyramid} & ICCV2019 & - & - & 78.3 & 68.0 \\
\multicolumn{1}{c|}{} &  & HPNet~\cite{huang2020human} & ICME2020 & - & - & \bftab{87.3} & \bftab{77.4} \\
\multicolumn{1}{c|}{} &  & DAReID~\cite{xu2021poseguide} & KBS2021 & \bftab{63.4} & \bftab{53.2} & - & - \\ \cline{2-8}

\multicolumn{1}{c|}{} & \multirow{6}{*}{\begin{tabular}[c]{@{}c@{}}Attention\\ Mechanism\\ for
Noise\end{tabular}} & IGOAS~\cite{wang2021occluded} & TIP2021 & 60.1 & 49.4 & 81.1 & - \\ 
\multicolumn{1}{c|}{} & & OAMN~\cite{chen2021occlude} & ICCV2021 & 62.6 & 46.1 & - & - \\
\multicolumn{1}{c|}{} &  & OCNet~\cite{kim2022ocnet} & ICASSP2022 & 64.3 & 54.4 & - & - \\
\multicolumn{1}{c|}{} &  & DRL-Net~\cite{jia2022learning} & TMM2021 & 65.8 & 53.9 & - & - \\
\multicolumn{1}{c|}{} &  & SSGR~\cite{yan2021occluded} & ICCV2021 & 65.8 & \bftab{57.2} & 78.5 & 72.9 \\
\multicolumn{1}{c|}{} &  & FED~\cite{wang2022fed} & CVPR2022 & \bftab{68.1} & 56.4 & \bftab{86.3} & \bftab{79.3} \\ \hline

\multicolumn{1}{c|}{\multirow{2}{*}{\begin{tabular}[c]{@{}c@{}}Missing\\ Information\end{tabular}}} & Spatial Recovery & RFCNet~\cite{hou2021completion} & TPAMI2021 & \bftab{63.9} & \bftab{54.5} & - & - \\ \cline{2-8}
\multicolumn{1}{c|}{} & Temporal Recovery & RFCNet~\cite{hou2021completion} & TPAMI2021 & \bftab{63.9} & \bftab{54.5} & - & - \\ \bottomrule
\end{tabular}}
\end{spacing}
\vspace{-0.5cm}
\end{table}

\section{Conclusion}
This paper aims at providing a systematic survey of occluded person Re-ID to help promote future research.
We first analyze and summarize four issues brought by occlusion in person Re-ID: the position misalignment, the scale misalignment, the noisy information, and the missing information.
The published publications of deep learning-based occluded person Re-ID from top conferences and journals before June, 2022 are categorized and introduced accordingly.
We provide the performance comparison of recent occluded person Re-ID methods on four popular datasets: Partial-ReID, Partial-iLIDS, Occluded-ReID, and Occluded-DukeMTMC.
Based on the analysis of evaluation results, we finally discuss the promising future research directions.
%

% {\small
% \bibliographystyle{IEEEtran}
% \bibliography{mybibfile}
% }

{\small
\bibliographystyle{unsrt2authabbrvpp}
\bibliography{manuscript}
}

\vspace{-26pt}
\begin{IEEEbiographynophoto}{Yunjie Peng}
received her B.S. degree in the College of Computer and Information Science \& College of Software from Southwest University, China, in 2018.
She is currently a Ph.D. student in the School of Computer Science and Technology, Beihang University, China.
Her research interests include gait recognition, perosn re-identification, computer vision, and machine learning.
\end{IEEEbiographynophoto}

% \begin{IEEEbiography}[{\includegraphics[width=1in,height=1.25in,clip,keepaspectratio]{bio_yunjiepeng}}]{Yunjie Peng}
% received her B.S. degree in the College of Computer and Information Science \& College of Software from Southwest University, China, in 2018.
% %
% She is currently a Ph.D. student in the School of Computer Science and Technology, Beihang University, China.
% %
% Her research interests include gait recognition, perosn re-identification, computer vision, and machine learning.
% \end{IEEEbiography}

\vspace{-26pt}
\begin{IEEEbiographynophoto}{Saihui Hou}
received the B.E. and Ph.D. degrees from University of Science and Technology of China in 2014 and 2019, respectively.
He is currently an Assistant Professor with School of Artificial Intelligence, Beijing Normal University.
His research interests include computer vision and machine learning.
He recently focuses on gait recognition which aims to identify different people according to the walking patterns.
\end{IEEEbiographynophoto}

% \begin{IEEEbiography}[{\includegraphics[width=1in,height=1.25in,clip,keepaspectratio]{bio_saihuihou}}]{Saihui Hou}
% received the B.E. and Ph.D. degrees from University of Science and Technology of China in 2014 and 2019, respectively.
% %
% He is currently an Assistant Professor with School of Artificial Intelligence, Beijing Normal University.
% %
% His research interests include computer vision and machine learning.
% %
% He recently focuses on gait recognition which aims to identify different people according to the walking patterns.
% \end{IEEEbiography}
% \vspace{11 pt}

\vspace{-26pt}
\begin{IEEEbiographynophoto}{Chunshui Cao}
received the B.E. and Ph.D. degrees from University of Science and Technology of China in 2013 and 2018, respectively.
During his Ph.D. study, he joined Center for Research on Intelligent Perception and Computing, National Laboratory of Pattern Recognition, Institute of Automation, Chinese Academy of Sciences.
From 2018 to 2020, he worked as a Postdoctoral Fellow with PBC School of Finance, Tsinghua University.
He is currently a Research Scientist with Watrix Technology Limited Co. Ltd. 
His research interests include pattern recognition, computer vision and machine learning.
\end{IEEEbiographynophoto}

% \begin{IEEEbiography}[{\includegraphics[width=1in,height=1.25in,clip,keepaspectratio]{bio_chunshuicao}}]{Chunshui Cao}
% received the B.E. and Ph.D. degrees from University of Science and Technology of China in 2013 and 2018, respectively.
% %
% During his Ph.D. study, he joined Center for Research on Intelligent Perception and Computing, National Laboratory of Pattern Recognition, Institute of Automation, Chinese Academy of Sciences.
% %
% From 2018 to 2020, he worked as a Postdoctoral Fellow with PBC School of Finance, Tsinghua University.
% %
% He is currently a Research Scientist with Watrix Technology Limited Co. Ltd. 
% %
% His research interests include pattern recognition, computer vision and machine learning.
% \end{IEEEbiography}

\vspace{-26pt}
\begin{IEEEbiographynophoto}{Xu Liu}
received the B.S. and Ph.D. degrees in control science and engineering from the University of Science and Technology of China (USTC), Hefei, China, in 2013 and 2018, respectively.
He is currently an algorithm researcher at Watrix Technology Limited Co. Ltd.
His current research interests include gait recognition, object detection, segmentation and deep learning.
\end{IEEEbiographynophoto}

% \begin{IEEEbiography}[{\includegraphics[width=1in,height=1.25in,clip,keepaspectratio]{bio_xuliu}}]{Xu Liu}
% received the B.S. and Ph.D. degrees in control science and engineering from the University of Science and Technology of China (USTC), Hefei, China, in 2013 and 2018, respectively.
% %
% He is currently an algorithm researcher at Watrix Technology Limited Co. Ltd.
% %
% His current research interests include gait recognition, object detection, segmentation and deep learning.
% \end{IEEEbiography}

\vspace{-26pt}
\begin{IEEEbiographynophoto}{Yongzhen Huang}
received the B.E. degree from Huazhong University of Science and Technology in 2006, and the Ph.D. degree from Institute of Automation, Chinese Academy of Sciences in 2011.
He is currently an Associate Professor with School of Artificial Intelligence, Beijing Normal University.
He has published one book and more than 80 papers at international journals and conferences such as TPAMI, IJCV, TIP, TSMCB, TMM, TCSVT, CVPR, ICCV, ECCV, NIPS, AAAI.
His research interests include pattern recognition, computer vision and machine learning.
\end{IEEEbiographynophoto}

% \begin{IEEEbiography}[{\includegraphics[width=1in,height=1.25in,clip,keepaspectratio]{bio_yongzhenhuang}}]{Yongzhen Huang}
% received the B.E. degree from Huazhong University of Science and Technology in 2006, and the Ph.D. degree from Institute of Automation, Chinese Academy of Sciences in 2011.
% %
% He is currently an Associate Professor with School of Artificial Intelligence, Beijing Normal University.
% %
% He has published one book and more than 80 papers at international journals and conferences such as TPAMI, IJCV, TIP, TSMCB, TMM, TCSVT, CVPR, ICCV, ECCV, NIPS, AAAI.
% %
% His research interests include pattern recognition, computer vision and machine learning.
% \end{IEEEbiography}

\vspace{-26pt}
\begin{IEEEbiographynophoto}{Zhiqiang He}
is currently the Senior Vice President of Lenovo Company and President of Lenovo Capital and Incubator Group.
This group is responsible for exploring external innovation as well as accelerating internal innovation for Lenovo Group, leveraging Lenovo global resources, power of capital, and entrepreneurship.
Previously, he was the Chief Technology Officer and held various leadership positions in Lenovo, particularly in overseeing Lenovoâs Research \& Technology initiatives and systems.
He is a doctoral supervisor at the Institute of Computing Technology, Chinese Academy of Sciences and Beihang University.
\end{IEEEbiographynophoto}

% \begin{IEEEbiography}[{\includegraphics[width=1in,height=1.25in,clip,keepaspectratio]{bio_zhiqianghe}}]{Zhiqiang He}
% is currently the Senior Vice President of Lenovo Company and President of Lenovo Capital and Incubator Group.
% %
% This group is responsible for exploring external innovation as well as accelerating internal innovation for Lenovo Group, leveraging Lenovo global resources, power of capital, and entrepreneurship.
% %
% Previously, he was the Chief Technology Officer and held various leadership positions in Lenovo, particularly in overseeing Lenovoâs Research \& Technology initiatives and systems.
% %
% He is a doctoral supervisor at the Institute of Computing Technology, Chinese Academy of Sciences and Beihang University.
% \end{IEEEbiography}

% \vspace{140 mm}
\vfill

\end{document}